\documentclass{article}

\PassOptionsToPackage{numbers, compress}{natbib}
\usepackage[preprint]{neurips_2026}

\usepackage[utf8]{inputenc} 
\usepackage[T1]{fontenc}    
\usepackage{url}            
\usepackage{booktabs}       
\usepackage{amsfonts}       
\usepackage{nicefrac}       
\usepackage{microtype}      
\usepackage{xcolor}         
\usepackage{amsmath, amsthm}
\usepackage{amssymb}
\usepackage{mathtools}
\usepackage[american]{babel}
\usepackage{tikz}
\usepackage{graphicx}
\usepackage{subcaption}
\usepackage{hyperref}
\usepackage{cleveref}

\newcommand{\eg}{\textit{e.g.}}
\newcommand{\ie}{\textit{i.e.}}
\usepackage{algorithm}
\usepackage{algorithmic}
\usepackage{rotating}
\usepackage{multirow}
\usepackage{todonotes}
\usepackage{wrapfig}
\usepackage{float}
\title{Solving Integer Linear Programming with \\  Parallel Tempering}

\author{%
  Kyuil Sim$^1$\quad
  Sanghyeok Choi$^2$\thanks{Equal advising. Correspondence to: \texttt{kyuil.sim@kaist.ac.kr}.}\quad
  Jinkyoo Park$^{1*}$\\[2mm]
  \textsuperscript{1}KAIST\quad
  \textsuperscript{2}University of Edinburgh
}

\begin{document}

\maketitle
\setcounter{footnote}{0}

\begin{abstract}
Integer Linear Programming (ILP) serves as a versatile framework for modeling a wide range of combinatorial optimization problems, typically addressed by sophisticated exact solvers or heuristics. While learning-based approaches have recently shown their effectiveness, they suffer from poor generalization to out-of-distribution instances and inherent dependence on external solvers. In this work, we propose a solver-free, sampling-based optimization framework for ILP that directly explores discrete feasible regions without training or external solvers. Exploiting the linear structure of ILP, we employ a Locally-Balanced Proposal to construct a transition kernel, thereby avoiding the gradient approximation. To overcome the highly multimodal nature of ILP energy landscapes, we integrate Parallel Tempering. In addition to standard temperature tempering, we introduce penalty tempering, which modulates constraint barriers while preserving the objective landscape over feasible solutions. Empirically, our method consistently outperforms SCIP across all four benchmarks, matches or exceeds Gurobi on two of four tasks within a 200-second budget, and is substantially more robust to distribution shift than learning-based methods. Furthermore, on MIPLIB 2017 instances, our framework remains competitive with classical solvers without any problem-specific tuning.\footnote{Source code is available at: \url{https://github.com/ski-sim/ILP-with-ParallelTempering}.}
\end{abstract}

\section{Introduction}\label{sec:introduction}

Integer Linear Programming (ILP) serves as a versatile framework for modeling a wide range of combinatorial optimization problems, including planning \citep{beyer2016solving, schuster2020exact}, scheduling \citep{ryan1981integer, trilling2006nurse}, vehicle routing \citep{kulkarni1985integer}, and portfolio optimization \citep{benati2007mixed}. Due to the combinatorial nature and NP-hardness of ILP, solving ILPs remains a significant challenge. The classical solvers, such as SCIP \citep{achterberg2009scip} and Gurobi \citep{optimization2020gurobi}, primarily build upon classical techniques, especially Branch-and-Bound (B\&B) \citep{land2009automatic} and Branch-and-Cut (B\&C) \citep{mitchell2002branch}. Although they are guaranteed to find the optimal solution given an infinite amount of time, the scalability of these solvers relies on sophisticated heuristics, which makes them harder to use.

Recently, leveraging machine learning (ML) within heuristic frameworks has emerged as a promising approach to accelerate optimization \citep{zhang2023survey, li2024machine}. One line of work attempts to replace components of B\&B \citep{gasse2019exact, feng2025sorrel, vo2025learning, kuang2025accelerate} or Large Neighborhood Search \citep{song2020general, wu2021learning, liu2022learning, huang2023searching, kong2024ilp, ye2025large, yuan2024btbs} with learned models. Another line of research trains models to directly predict feasible solutions, thereby providing high-quality initial solutions for classical or heuristic solvers \citep{nair2020solving,yoon2022confidence, han2023gnn, huang2024contrastive, zeng2024effective, geng2025differentiable}.

Despite these advances, learning-based approaches suffer from two fundamental limitations: poor generalization to out-of-distribution instances and inherent dependence on external solvers.

In this work, we explore sampling-based approaches \citep{vcerny1985thermodynamical, neal1996sampling, secckiner2007simulated, wong2012simulated, ma2019sampling, dong2021replica} as an alternative direction to overcome these limitations, inspired by recent works  \citep{sun2023revisiting,feng2025regularized, wang2026fractional} that have demonstrated the effectiveness of Markov Chain Monte Carlo (MCMC) approaches for combinatorial optimization (CO). While these works for CO usually rely on the gradient of the target density to approximate the Locally Balanced Proposal \citep[LBP;][]{zanella2020informed}, applying them directly to ILP is problematic because the inherent linear structure of ILP makes gradient information uninformative.

We address these challenges through an algorithmic design tailored to the linear structure of ILP. From the observation that the exact Locally Balanced Proposal \citep[LBP;][]{zanella2020informed} can be obtained efficiently thanks to the linear structure of ILP, we use a simple yet effective gradient-free multi-step proposal that approximates LBP.
While this enables efficient local moves, ILP energy landscapes remain highly multimodal and fragmented by constraints; we therefore use parallel tempering (PT; also known as replica exchange) as a global exploration mechanism. Specifically, we investigate two distinct PT strategies, each tempering a different parameter: temperature $\tau$ and penalty $\lambda$. While classical $\tau$-based PT alters the entire energy surface, $\lambda$-based PT selectively modifies the barrier in infeasible regions while preserving the energy landscape in feasible regions.

We evaluate our framework on standard ILP benchmarks with four different classes of problems, under both in-distribution and out-of-distribution settings. Our sampling-based approach is competitive with Gurobi on two out of four tasks within a 200-second budget, consistently outperforms SCIP, and shows better robustness under distribution shifts compared to learning-based methods. Beyond synthetic benchmarks, our framework matches the performance of classical solvers on real-world MIPLIB 2017 instances without problem-specific tuning.

\begin{figure*}[t]
    \centering
    \includegraphics[width=1.0\linewidth]{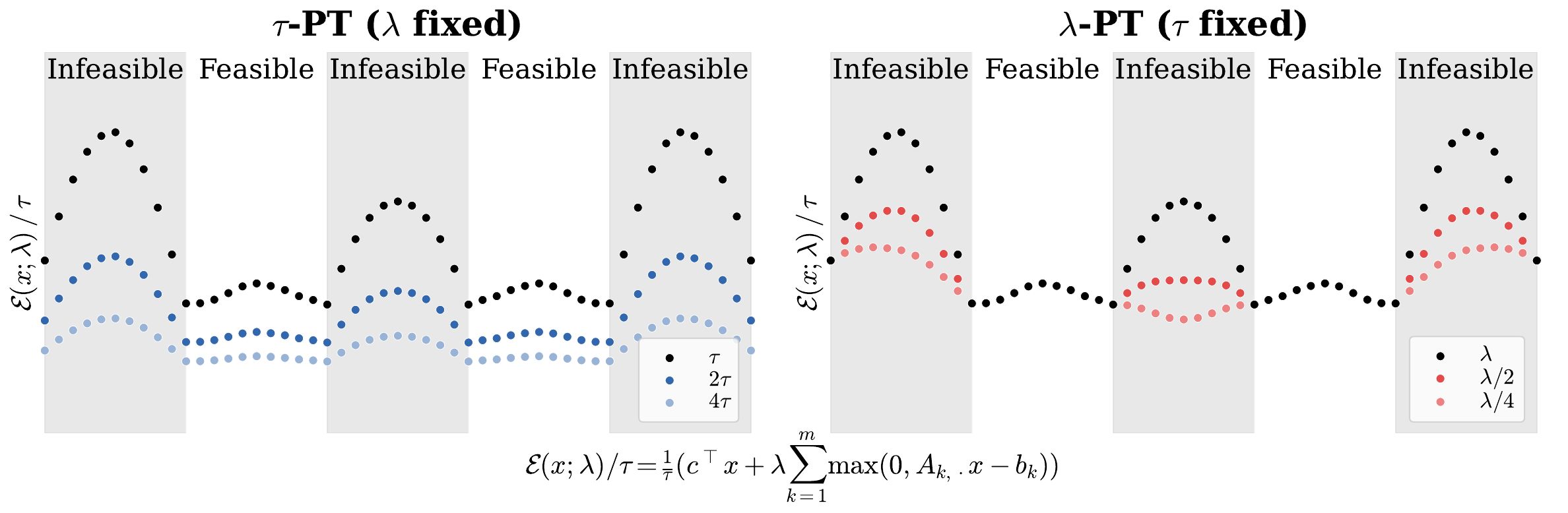}
    \caption{Visualization of two parallel tempering (PT) strategies for discrete sampling.
    \textcolor{blue}{\textbf{(Left) $\tau$-PT}}: 
    Higher temperatures (\eg, $4\tau$) flatten the entire energy surface, facilitating exploration across both feasible and infeasible regions. \textcolor{red}{\textbf{(Right) $\lambda$-PT}}: 
    Lower penalty weights (\eg, $\lambda/4$) selectively reduce the barrier in infeasible regions while preserving the energy landscape in feasible regions.}
    \label{fig:pt-comparison}
    \vspace*{-1em}
\end{figure*}
\section{Background}\label{sec:background}
\subsection{Integer linear programming}
A canonical form of ILP with binary variables is formulated as
\begin{equation}
  \label{eq:ilp}
  \min_x \space  c^{\top}x \quad  \text{s.t} \quad   A x \leq b \quad \text{and} \quad  x\in \{0,1\}^n,
\end{equation}
where $n$ and $m$ denote the number of variables and constraints, respectively, with $c \in \mathbb{R}^n, A \in \mathbb{R}^{m \times n}, b \in \mathbb{R}^{m}$. Extending to general integer variables is straightforward. Due to the NP-hard nature of ILP \citep{karp2009reducibility}, researchers often employ heuristics such as Large Neighborhood Search \citep{pisinger2018large} or Local Branching \citep{fischetti2003local} to find high-quality feasible solutions quickly.
We summarize recent advances in ILP, especially machine-learning-based approaches, in \Cref{sec:realtedworks}.

\subsection{Metropolis-hastings algorithm}
Given a target distribution $\pi(x)$, the Metropolis-Hastings \citep[MH;][]{metropolis1953equation,hastings1970monte} algorithm generates a Markov chain where each transition at time $t$ consists of two stages: 1) propose a candidate state $y$ from a \emph{proposal} distribution $q(y \mid x)$, and 2) accept it with probability
\begin{equation}
\alpha(y,x):=\min\left(1, \frac{\pi(y)q(x \mid y)}{\pi(x)q(y \mid x)}\right),
\end{equation}
\ie, replace $x$ by $y$ with probability $\alpha(y,x)$. Assuming the proposal guarantees ergodicity, the algorithm satisfies detailed balance, and the Markov chain has $\pi(x)$ as its invariant distribution \citep{bishop2006pattern}.

\section{Solving ILP via sampling}

\subsection{Formulation}
\label{sec:formulation}

In this subsection, we cast the ILP as a sampling problem and discuss the form of the target distribution and proposal distributions for running the MH algorithm in discrete sample spaces.

\subsubsection{Target distribution}
\label{sec:target}
We reformulate the ILP \eqref{eq:ilp} by integrating the objective function and constraints into a single target distribution $\pi(x)$. In this work, we use the form of the Gibbs distribution whose energy function is defined as 
\begin{equation}
\label{eq:energy}
  \mathcal{E}(x;\lambda)  \coloneq c^{\top}x + \lambda \sum_{k=1}^{m}\max(0, A_{k,\cdot} x - b_k),
\end{equation}
where $A_{k,\cdot}$ is the $k$-th row of matrix $A$, and $\lambda$ is the penalty parameter to encourage constraint satisfaction. This formulation is natural given the inequality constraints commonly found in ILP. Note that this formulation is distinct from previous works \citep{sun2023revisiting,feng2025regularized} that used the standard Lagrangian form for CO problems with only equality constraints. Converting an ILP with inequality constraints into one with equality constraints introduces a slack variable for each constraint, which is undesirable in sampling settings. The target distribution is then defined as:
\begin{equation}
\label{eq:ilp_sampling}
  \pi(x;\lambda,\tau) = \frac {\exp(-\mathcal{E}(x;\lambda)/\tau)}{Z(\lambda,\tau)},
\end{equation}
where $\tau$ is a temperature parameter and $Z(\lambda,\tau) = \sum_{x \in \{0,1\}^{n}} \exp(-\mathcal{E}(x;\lambda)/\tau)$.

Since the goal is optimization, our approach is based on the \emph{simulated annealing} \citep[SA;][]{kirkpatrick1983optimization,johnson1991optimization}, in which the temperature $\tau$ is scheduled to decrease throughout the sampling procedure. Specifically, we set $\tau$ as a function of the time step $t$, \eg, exponentially decaying $\tau(t)=\tau_0\times\gamma^t$ where $\gamma$ is the decay rate and $\tau_0$ is the initial temperature. This schedule follows $\tau(0) > \tau(1) > \dots > \tau(T) \to 0$, allowing for a stable transition toward the global optimum of $\mathcal{E}(\cdot;\lambda)$  as the iteration count $T$ increases. Note also that we must choose a sufficiently large $\lambda$ such that $x^{\star} = \arg\min \mathcal{E}(x;\lambda)$ solves \eqref{eq:ilp}, although $\lambda$ is not necessarily scheduled toward $\infty$ as it only affects infeasible solutions.

\subsubsection{Proposal distribution}
\label{sec:proposal}
The design of the transition kernel is critical to the mixing rate of the MH algorithm. Conventional approaches such as random-walk proposals and Gibbs sampling often suffer from slow convergence in complex, structured target spaces. In this work, we focus on the \emph{locally-balanced proposal} \citep[LBP;][]{zanella2020informed}, a family of proposals with asymptotic optimality in discrete spaces.

We first introduce the pointwise informed proposal (PIP), which is defined as
\begin{equation}
\label{eq:pip}
    q^{\mathcal{N}}_g \left(y \mid x\right) \propto g\left(\frac{\pi(y;\lambda,\tau)}{\pi(x;\lambda,\tau)}\right)1(y\in \mathcal{N}(x))
\end{equation}
where $\mathcal{N}(x)$ is a neighborhood around $x$ and $g: [0,\infty) \to [0, \infty]$ is a continuous function. LBP is a special case of PIP, which uses the locally balancing function $g$ that satisfies $g(t) = t g(1/t)$ for all $t\in[0, \infty)$. In this work, we fix $g(t)=\sqrt{t}$. When regularity assumptions hold in the high-dimensional regime, LBP is asymptotically optimal in terms of Peskun ordering \citep{peskun1973optimum}.\footnote{Peskun ordering compares Markov chain convergence, using asymptotic variance and spectral gap to measure MCMC efficiency. See Section 3 of \citet{zanella2020informed} for details}.

Evaluating the unnormalized density for all neighbors $y \in \mathcal{N}(x)$ is often prohibitively expensive to compute. To circumvent this, several studies approximate LBP by first-order Taylor expansions \citep{grathwohl2021oops, sun2021path, zhang2022langevin} using the gradient of the target density. Recent works in combinatorial optimization have also adopted these gradient-based approximations of LBP \citep{sun2023revisiting, feng2025regularized}.

A key advantage of the ILP setting is that LBP can be obtained exactly and efficiently, thanks to the linear structure of the energy function \eqref{eq:energy}. For a given $x\in \{0,1\}^n$, let $x^{-j}$ be a vector obtained by flipping the $j$-th coordinate of $x$, and $\mathcal{N}_{1}(x)=\{x^{-1},x^{-2},\ldots,x^{-n}\}$ be the 1-Hamming sphere around $x$.\footnote{The $r$-Hamming sphere is the set of points at Hamming distance exactly $r$ from a center. Note the distinction from the $r$-Hamming ball, which is the set of points within distance $r$.} Note also that $c^{\top}x^{-j}$ and $Ax^{-j}$ can be expressed as follows:
\begin{equation*}
c^{\top}x^{-j} = c^{\top}x + c_{j}(1-2x_{j}), \quad Ax^{-j} = Ax + (1-2x_j) \cdot A_{\cdot,j}.
\end{equation*}
We can efficiently compute both of them, for all $j \in 1, \ldots ,n$, by simple matrix multiplications:
\begin{align*}
\left[c^{\top}x^{-1} \cdots c^{\top}x^{-n} \right] &= c^{\top}x \cdot \mathbf{1}_n^{\top} + c^\top \odot (1-2x^\top), \!\!\!\\
\begin{bmatrix}
\vert & & \vert \\
Ax^{-1} & \!\!\!\cdots\!\!\! & Ax^{-n} \\
\vert & & \vert
\end{bmatrix} &= Ax \cdot \mathbf{1}_{n}^{\top} + A \cdot \operatorname{diag}(1-2x),\!\!\!
\end{align*}
where $\odot$ is element-wise multiplication. With these, the energy values can be computed using \eqref{eq:energy}, which in turn yields the LBP within the 1-Hamming sphere by
\begin{equation}
    \label{eq:lbp-1hamming}
    q^{\mathcal{N}_1}_g(x^{-j} \mid x) \propto g\left( \exp(-\mathcal{E}(x^{-j}) + \mathcal{E}(x)) \right)
\end{equation}

However, moving within the 1-Hamming sphere at a time can be inefficient in practice. Thus, we propose sampling $L$ indices \emph{without replacement} from \eqref{eq:lbp-1hamming}, which allows us to take larger MH steps. We refer to the resulting proposal as \textbf{multi-step LBP} (MLBP) with MLBP-$L$ when $L$ indices are updated. Although the theoretical optimality of the exact LBP no longer holds, MLBP provides a more calibrated approximation to the LBP than the aforementioned gradient-based approaches, especially because the gradient of \eqref{eq:energy} is piecewise constant.

\subsection{Parallel tempering}

The energy landscape of ILP is inherently rugged and multimodal due to the discrete, combinatorial nature of the solution space. Vanilla MCMCs often become trapped in local optima, unable to escape narrow feasible regions separated by high-energy barriers. Parallel Tempering (PT) addresses this by maintaining multiple Markov chains simultaneously, each with a different temperature $\tau$: high-temperature chains traverse steep energy barriers to facilitate global exploration, while low-temperature chains refine solutions within promising regions. 
Although PT has been extensively studied in continuous domains, we adapt PT to ILP sampling, enabling chains to discover diverse modes of high-quality feasible solutions.

We explore two PT strategies: the standard $\tau$-PT, which tempers the 
temperature $\tau$, and our proposed $\lambda$-PT, which tempers the penalty 
parameter $\lambda$ to selectively modulate constraint barriers. These two mechanisms differ fundamentally in how they facilitate inter-region transitions, as illustrated in \Cref{fig:pt-comparison}, which contrasts their effects on the energy landscape.

\subsubsection{$\tau$-parallel tempering}
\label{sec:temperature parallel tempering}
$\tau$-PT runs $B$ chains in parallel, each targeting a tempered distribution at a distinct temperature $\tau^1 < \tau^2 < \cdots < \tau^B$ with a fixed penalty $\lambda$. The joint state $\mathbf{x} = (x^1, \ldots, x^B) \in \mathcal{X}^B$ is maintained with invariant joint distribution:
\begin{equation}
\pi(\mathbf{x};\lambda,\tau^1,\ldots,\tau^B) = \prod_{i=1}^{B} \pi(x^i;\lambda,\tau^i).
\end{equation}
To promote inter-chain mixing, PT periodically proposes state exchanges between adjacent chains. Given states $x^i$ and $x^j$ at adjacent temperatures $\tau^i < \tau^j$, the swap is accepted with probability:
\begin{equation}
\label{temperature_swap}
\alpha^\tau_{swap}(x^i,x^j) =\min\!\left(1,\ \frac{\pi(x^i;\lambda,\tau^j)\pi(x^j;\lambda,\tau^i)}{\pi(x^i;\lambda,\tau^i)\pi(x^j;\lambda,\tau^j)}\right)= \min\!\left(1,\ \exp\!\left(\Delta\beta_{i,j} \cdot \Delta\mathcal{E}_{i,j}\right)\right)
\end{equation}
where $\Delta\beta_{i,j} = \frac{1}{\tau^i} - \frac{1}{\tau^j}$ and $\Delta\mathcal{E}_{i,j} = \mathcal{E}(x^i;\lambda) - \mathcal{E}(x^j;\lambda)$.
To further improve mixing—measured by the round-trip rate—we employ the non-reversible PT scheme \citep{syed2022non}; details are provided in Appendix~\ref{app:non-reversible}. The temperature ladder is initialized with exponential spacing $\tau^1_{(0)} = \tau_{\min}, \ldots, \tau^B_{(0)} = \tau_{\max}$, and each chain's temperature decays as $\tau^i_{(t)} = \tau^i_{(0)} \times \gamma^{t}$.

\subsubsection{$\lambda$-parallel tempering}
\label{sec:penalty parallel tempering}
While $\tau$-PT modulates the entire energy landscape, the ILP landscape is primarily shaped by constraint barriers controlled by the penalty parameter $\lambda$. High penalty values create sharp discontinuities at constraint boundaries, making it difficult for chains to transition between feasible and infeasible regions. In this regard, we propose $\lambda$-PT that varies $\lambda$ across chains at a fixed temperature $\tau$, selectively 
reshaping these constraint barriers.

As before, $\lambda$-PT maintains $B$ chains, each targeting $\pi(x;\lambda^i,\tau)$ with a distinct penalty level $\lambda^1 > \lambda^2 > \cdots > \lambda^B$. The invariant joint distribution is:
\begin{equation}
\pi(\mathbf{x};\tau,\lambda^1,\ldots,\lambda^B) = \prod_{i=1}^{B} \pi(x^i;\lambda^i,\tau).
\end{equation}
Applying the detailed balance condition to this joint distribution, the swap acceptance probability between adjacent chains $i$ and $j$ with $\lambda^i < \lambda^j$ is:
\begin{equation}
\label{penalty_swap}
\alpha^\lambda_{swap}(x^i,x^j) =\min\!\left(1,\ \frac{\pi(x^i;\lambda^j,\tau)\pi(x^j;\lambda^i,\tau)}{\pi(x^i;\lambda^i,\tau)\pi(x^j;\lambda^j,\tau)}\right)= \min\!\left(1,\ \exp\!\left(\frac{\Delta\lambda_{i,j} \cdot \Delta P_{i,j}}{\tau}\right)\right)
\end{equation}
where $P(x) = \sum_{k=1}^{m}\max(0, A_{k,\cdot}x - b_k)$ is the total constraint violation, $\Delta\lambda_{i,j} = \lambda^j - \lambda^i$, and $\Delta P_{i,j} = P(x^j) - P(x^i)$.
Chains with lower $\lambda$ can freely traverse constraint boundaries to explore infeasible regions, while chains with higher $\lambda$ focus on feasible solutions. As in $\tau$-PT, the shared temperature also follows an exponential decay schedule $\tau_{(t)} = \tau_{(0)} \times \gamma^{t}$ so that all chains progressively concentrate on high-quality feasible solutions over time. 
Intuitively, $\lambda$-PT offers a parallel alternative to adaptive penalty 
schedules. Because the choice of $\lambda$ critically affects performance 
in constrained sampling, prior work has often relied on online adaptation 
of $\lambda$. $\lambda$-PT achieves a comparable effect by 
maintaining a ladder of $\lambda$ values across chains. The full algorithm with MLBP and both PT strategies is presented in Appendix~\ref{app:algorithms}.

\section{Related work}\label{sec:realtedworks}
\subsection{Recent approaches for ILP}
\textbf{Heuristic approaches.}
Solving ILPs remains a significant challenge due to the combinatorial nature and NP-hardness. Standard solvers such as SCIP \citep{achterberg2009scip} and Gurobi \citep{optimization2020gurobi} utilize Branch-and-Bound (B\&B) \citep{land2009automatic} enhanced by sophisticated heuristics. Large Neighborhood Search (LNS) \citep{pisinger2018large} serves as a powerful alternative by iteratively exploring large sub-regions of the search space.

\textbf{Learning-based approaches.}
Recently, learning-based approaches have been proposed to further accelerate both exact and heuristic algorithms. Most existing studies replace heuristic components with machine learning models. In the B\&B framework, learning has been applied to enhance key components such as branching decisions \citep{balcan2018learning, gasse2019exact, kuang2024rethinking, kuang2024towards, feng2025sorrel}, cut selection \citep{paulus2022learning, wang2023learning}, cut generation \citep{vo2025learning}, and presolving \citep{liu2024l2p, kuang2025accelerate}. In the LNS framework, learning is primarily applied to the destroy operator that selects variables to re-optimize, while external solvers handle the repair step \citep{hottung2019neural, ye2025large}; representative methods leverage imitation \citep{song2020general}, reinforcement \citep{wu2021learning}, and contrastive learning \citep{huang2023searching}, with further extensions to sequence-based formulations and binarized tightening \citep{kong2024ilp, yuan2024btbs}.

Another line of research uses ML to generate initial feasible solutions \citep{nair2020solving, yoon2022confidence, liu2025apollo}. PaS \citep{han2023gnn} and ConPaS \citep{huang2024contrastive} predict partial solutions and search within a trust region around the prediction rather than hard-fixing variables, while \citet{zeng2024effective} produce complete solutions via diffusion models. DiffILO \citep{geng2025differentiable} reformulates ILP-solving as a differentiable optimization problem, avoiding expensive label generation, though its scalability remains limited and it still depends on external solvers. By contrast, we propose a sampling-based approach that requires neither learning nor external solvers.

\subsection{Sampling-based approach for discrete optimization}
Sampling-based methods have a long history in combinatorial optimization (CO). Simulated Annealing \citep[SA;][]{kirkpatrick1983optimization, johnson1991optimization} is the classical example but its convergence typically necessitates extensive sampling. Recent advances in discrete MCMC have demonstrated the potential to accelerate convergence by employing locally balanced proposals \citep{zanella2020informed, sun2021path, zhang2022langevin}. Drawing on these advances together with theoretical insights into sampling-based optimization \citep{ma2019sampling, dong2021replica}, iSCO \citep{sun2023revisiting} reformulates CO problems using a Lagrangian-relaxed target distribution and employs the Path Auxiliary Fast Sampler proposal for sampling. 

A central remaining challenge is escaping local optima, which is more difficult in discrete than in continuous domains: ReSCO \citep{li2025reheated} addresses this through a reheating mechanism, while RLD \citep{feng2025regularized} incorporates expected Hamming distance regularization into the objective function. \citet{wang2026fractional} propose Fractional Langevin Dynamics, replacing the Gaussian noise in discrete Langevin dynamics with $\alpha$-stable Lévy noise to enable polynomial-time escape from local optima via Lévy flights.
Most closely related to our work, \citet{liang2025enhancing} apply PT with an adaptive temperature schedule to enhance exploration in discrete sampling. Building on these foundations, we develop a sampling-based framework specifically tailored to the linear structure of ILP.

\begin{table*}[t]
  \centering
  \caption{Evaluation of objective values and relative gaps against exact solvers (Gurobi) and heuristics across four ILP benchmarks within a 200-second time limit. Obj.\ denotes the mean objective value over 100 test instances, and Gap denotes mean $\pm$ standard deviation of the relative gap (\%).}
  \label{tab:main_results1}
  \resizebox{1.0\textwidth}{!}{
  \begin{tabular}{l cc cc cc cc cc cc cc cc cc}
    \toprule
    & \multicolumn{2}{c}{MVC-1000}
    & \multicolumn{2}{c}{MIS-1500}
    & \multicolumn{2}{c}{CA-2000}
    & \multicolumn{2}{c}{SC-2000}
    & \multicolumn{2}{c}{MVC-2000}
    & \multicolumn{2}{c}{MIS-3000}
    & \multicolumn{2}{c}{CA-4000}
    & \multicolumn{2}{c}{SC-4000}  \\
    \cmidrule(lr){2-3} \cmidrule(lr){4-5} \cmidrule(lr){6-7} \cmidrule(lr){8-9} \cmidrule(lr){10-11} \cmidrule(lr){12-13} \cmidrule(lr){14-15} \cmidrule(lr){16-17}
    Methods & Obj. $\downarrow$ & Gap $\downarrow$ & Obj. $\downarrow$ & Gap $\downarrow$ & Obj. $\downarrow$ & Gap $\downarrow$& Obj. $\downarrow$& Gap $\downarrow$& Obj. $\downarrow$& Gap $\downarrow$& Obj. $\downarrow$& Gap $\downarrow$ & Obj. $\downarrow$& Gap $\downarrow$& Obj. $\downarrow$ & Gap $\downarrow$\\
    \midrule
    Gurobi  & 444.4 & - & -684.5 & - & -65017 & - & 293.0 & - & 897.3 & - & -1372.3 & - & -128231 & - & 170.9 & -\\
    \midrule
    SCIP  & 460.9  &  3.71\scriptsize$\pm$0.64
          & -682.8 &  0.25\scriptsize$\pm$0.85
          & -63364 &2.54\scriptsize$\pm$0.84
          & 301.6 &  2.88\scriptsize$\pm$1.97
          & 919.8 &  2.50\scriptsize$\pm$0.50
          & -1362.2 & 0.74\scriptsize$\pm$0.60
          & -123964  & 3.32\scriptsize$\pm$1.03
          & 179.6 & 4.67\scriptsize$\pm$2.02\\
    DINS  & 460.0  &  3.52\scriptsize$\pm$0.75
          & -683.0 & 0.22\scriptsize$\pm$0.87
          & -63197 &2.80\scriptsize$\pm$0.95
          & 301.6 &  2.87\scriptsize$\pm$1.92
          & 919.8 &  2.50\scriptsize$\pm$0.50
          & -1362.3 & 0.74\scriptsize$\pm$0.60
          & -123964  & 3.32\scriptsize$\pm$1.03
          & 179.3  & 4.48\scriptsize$\pm$1.78\\
    GINS  & 459.4 & 3.38\scriptsize$\pm$0.79
          & -683.0 & 0.22\scriptsize$\pm$0.87
          & -63197 &2.80\scriptsize$\pm$0.95
          & 301.9 & 2.99\scriptsize$\pm$1.84
          & 919.8 &  2.50\scriptsize$\pm$0.50
          & -1362.3   & 0.74\scriptsize$\pm$0.60
          & -123964   & 3.32\scriptsize$\pm$1.03
          & 179.3 & 4.48\scriptsize$\pm$1.78\\
    RINS  & 459.3 &  3.35\scriptsize$\pm$0.79
          & -682.9 & 0.24\scriptsize$\pm$0.84
          & -63367  & 2.54\scriptsize$\pm$0.85
          & 300.7 &  2.57\scriptsize$\pm$2.00
          & 919.8 &  2.50\scriptsize$\pm$0.50
          & -1362.2 & 0.74\scriptsize$\pm$0.60
          & -123964  & 3.32\scriptsize$\pm$1.03
          & 179.6 & 4.67\scriptsize$\pm$2.02\\
    RENS  & 459.1 &  3.31\scriptsize$\pm$0.81
          & -683.0 &  0.22\scriptsize$\pm$0.87
          & -63197 & 2.80\scriptsize$\pm$0.95
          & 302.0 &  2.99\scriptsize$\pm$1.84
          & 919.8 &  2.50\scriptsize$\pm$0.50
          & -1362.3 & 0.74\scriptsize$\pm$0.60
          & -123944  &3.34\scriptsize$\pm$0.96
          & 179.3 & 4.48\scriptsize$\pm$1.78\\
    \midrule
    \textbf{MLBP}\\
    \textbf{\hspace{1pt} + SA } &\underline{442.5}  &\underline{-0.43\scriptsize$\pm$0.39}
                      & -683.6 & 0.14\scriptsize$\pm$0.13
                      & -63835 & 1.82\scriptsize$\pm$0.57
                      & {291.6}  & {-0.46\scriptsize$\pm$0.59}
                      & \underline{884.7}  &\underline{-1.41\scriptsize$\pm$0.41}
                      & \underline{-1369.1} &\underline{0.24\scriptsize$\pm$0.10}
                      & {-126118} &{1.64\scriptsize$\pm$0.81}
                      & \underline{170.6}&  \underline{-0.17\scriptsize$\pm$0.68} \\
    \textbf{\hspace{1pt} + SA + {\small Reheat}} & \textbf{442.5} & \textbf{-0.43\scriptsize$\pm$0.39}
                      &  \underline{-684.0}  & \underline{0.08\scriptsize$\pm$0.10}
                      & -63780 & 1.88\scriptsize$\pm$2.17
                      & \textbf{291.5} &  \textbf{-0.49\scriptsize$\pm$0.62}
                      &  884.7  & -1.41\scriptsize$\pm$0.41
                      &  -1369.1 &  0.24\scriptsize$\pm$0.10
                      & -126094   & 1.66\scriptsize$\pm$0.78
                      & 170.9 &  -0.01\scriptsize$\pm$0.70   \\
    \textbf{\hspace{1pt} + SA + $\tau$-PT} & 442.6 & -0.39\scriptsize$\pm$0.40
                                  & {-683.8} &{0.10\scriptsize$\pm$0.10}
                                  & \textbf{-63927}  & \textbf{1.65\scriptsize$\pm$2.21}
                                  & \underline{291.6}   & \underline{-0.47\scriptsize$\pm$0.60}
                                  & \textbf{883.6} &\textbf{-1.52\scriptsize$\pm$0.42}
                                  & -1369.0 & 0.25\scriptsize$\pm$0.11
                                  & \underline{-126212} & \underline{1.57\scriptsize$\pm$0.81}
                                  & 172.0 & 0.68\scriptsize$\pm$1.06 \\
    \textbf{\hspace{1pt} + SA + $\lambda$-PT} & {442.5} & {-0.41\scriptsize$\pm$0.40}
                                      &\textbf{-684.2} & \textbf{0.05\scriptsize$\pm$0.08}
                                      & \underline{-63887} & \underline{1.71\scriptsize$\pm$2.17}
                                      & 291.7  & -0.42\scriptsize$\pm$0.71
                                      & 885.3  & -1.34\scriptsize$\pm$0.43
                                      & \textbf{-1370.5} &  \textbf{0.13\scriptsize$\pm$0.08}
                                      & \textbf{-126283} &   \textbf{1.51\scriptsize$\pm$0.85}
                                      & \textbf{170.3}  & \textbf{-0.36\scriptsize$\pm$0.69} \\
    \bottomrule
  \end{tabular}
  }
\end{table*}
\begin{figure*}[t]
    \centering
    \includegraphics[width=\linewidth]{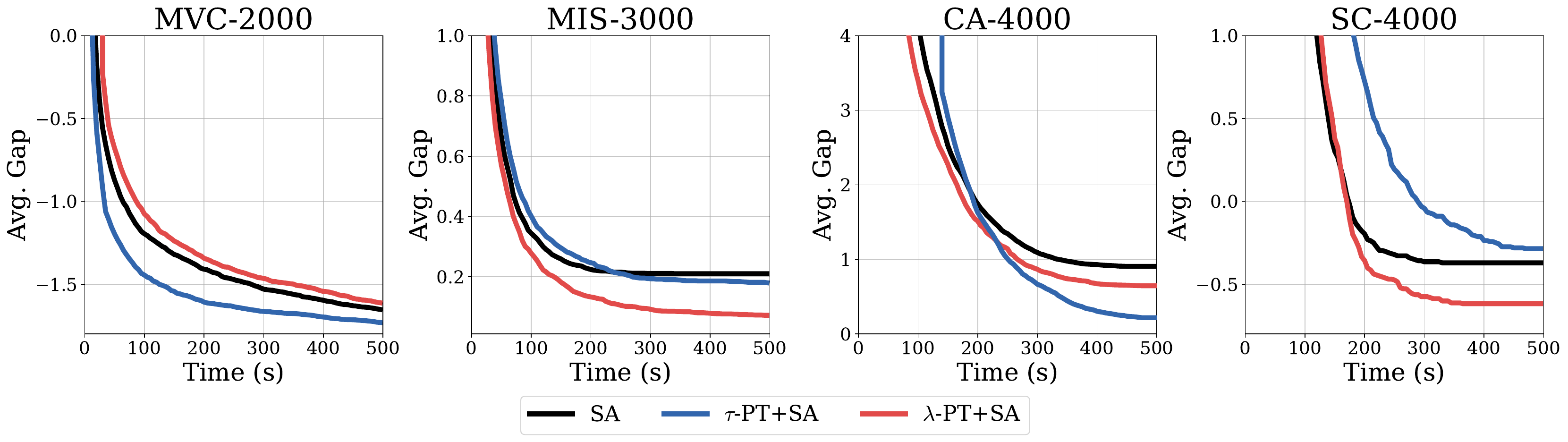}
    \caption{Any-time performance on MVC, MIS, CA, SC across different schedules under a 500-second budget.}
    \label{fig:runtime}
    \vspace{-15pt}
\end{figure*}

\section{Experiments}\label{sec:experiments}
This section empirically validates our sampling-based framework for ILP. \Cref{sec:overall} assesses overall performance and scalability against classical solvers and heuristics. \Cref{sec:OOD} examines robustness on out-of-distribution (OOD) instances and real-world MIPLIB benchmarks, in comparison to learning-based methods. \Cref{sec:additional analysis} provides additional analyses, including comparisons against alternative discrete samplers and an ablation of the simulated annealing schedule.

\textbf{Instances.} We focus on four NP-hard benchmarks following \citet{huang2023searching}: Minimum Vertex Cover (MVC), Maximum Independent Set (MIS), Combinatorial Auction (CA), and Set Covering (SC). For each problem, we generate 20 instances for validation and 100 for testing. We further evaluate our framework on two additional test sets: 100 double-sized instances to assess scalability, and 20 OOD instances—generated by shifting the underlying distribution parameters—to verify robustness under distribution shifts. The number of variables is indicated by a suffix (e.g., MIS-3000); detailed descriptions of the instance generation process are provided in Appendix~\ref{app:instance-generation}. 
 
\textbf{Hyperparameters.} The key hyperparameters are temperature $\tau$ and penalty parameter $\lambda$, which are determined via a small-scale grid search on the validation set; the detailed tuning procedure is provided in Appendix~\ref{app:validation-tuning}. The selected hyperparameters are fixed across all problem instances within the same problem class. For the proposal, we use MLBP-3 ($L=3$) throughout all experiments. Other design choices, including the swap interval $I$ and the number of chains, exhibit minimal impact on performance and are further discussed in Appendix~\ref{app:additional}.

\textbf{Evaluation metric.} The algorithms are evaluated based on (1) the objective value of incumbent solution attained within the time limit, and (2) the relative gap to the best-known solution ($BKS$). The relative gap is defined as $\frac{OBJ-BKS}{|BKS|} \times 100$, where $OBJ$ denotes the objective value of the incumbent solution found by a given method. We utilize results from Gurobi \citep{optimization2020gurobi} with multi-processing for $BKS$.

\textbf{Baselines.} We compare our method with classical solvers, including Gurobi~\citep{optimization2020gurobi},
SCIP~\citep{achterberg2009scip}, and heuristic methods, including RENS, RINS, DINS, GINS \citep{achterberg2007constraint}, and learning-based methods such as IL-LNS \citep{sonnerat2021learning}, CL-LNS \citep{huang2024contrastive}. For IL-LNS and CL-LNS, Both models were trained on a dataset consisting of 1,024 instances obtained from the Ecole library \citep{gasse2019exact}. Other recent LNS-based methods \citep{kong2024ilp, yuan2024btbs} were excluded because their source codes are unavailable. Detailed explanation of the baseline implementations is provided in~Appendix~\ref{app:baselines-details}. Our approach employs the LBP proposal with three distinct scheduling strategies: Simulated Annealing (SA), $\tau$-Parallel Tempering with Annealing ($\tau$-PT+SA), and $\lambda$-Parallel Tempering with Annealing ($\lambda$-PT+SA). We additionally include SA+reheat, a SA variant with periodic temperature reheating as a baseline. 

\subsection{Overall performance evaluation}
\label{sec:overall}
\textbf{Research question}: \textit{How does the proposed sampling framework perform compared to classical exact solvers like SCIP under strict runtime constraints (200s)?}

\paragraph{Main results.} As shown in the left half of \Cref{tab:main_results1}, our sampling-based methods consistently outperform SCIP across all benchmarks. Most notably, our approach surpasses Gurobi on both MVC and SC tasks within a strict 200-second runtime limit. These results position our methods as highly competitive, general-purpose solvers. 
As shown on the right side of \Cref{tab:main_results1}, our approach maintains stable, high-quality performance even when problem sizes are doubled. The results further indicate that $\lambda$-PT+SA begins to outperform other methods as the number of constraints increases, highlighting its superior handling of constraint-dense feasibility landscapes. In contrast, the reheat mechanism shows its strength, but its efficacy diminishes on doubled-size instances.

\paragraph{Extended runtime.}
Under a strict 200-second time budget, standard SA achieves performance comparable to PT methods (see \Cref{tab:main_results1}). We hypothesize that within such a short window, exploitation around a local optimum is highly advantageous, allowing SA to perform well. To verify whether the PT methods possess the ability to escape local optima, we extend the runtime to 500s. As shown in \Cref{fig:runtime}, PT methods indeed fall into local optima later than SA, allowing them to keep improving over longer runtimes. Notably, $\tau$-PT+SA yields larger gains on CA (which has relatively few constraints), whereas $\lambda$-PT+SA excels on MIS and SC (which feature dense, complex constraints).


\begin{table*}[t]
\centering
\caption{Generalization performance of learning-based baselines versus our sampling framework on OOD instances within a 200-second time limit.}
\label{tab:ood_results1}
\resizebox{1.0\textwidth}{!}{
\begin{tabular}{l cc cc cc cc cc cc cc cc cc}
    \toprule
& \multicolumn{8}{c}{\textbf{In-Distribution}} & \multicolumn{8}{c}{\textbf{Out-of-Distribution }} \\
    \cmidrule(lr){2-9} \cmidrule(lr){10-17}
  & \multicolumn{2}{c}{MVC-1000}
  & \multicolumn{2}{c}{MIS-1500}
  & \multicolumn{2}{c}{CA-2000}
  & \multicolumn{2}{c}{SC-2000}
  & \multicolumn{2}{c}{MVC-1000}
  & \multicolumn{2}{c}{MIS-1500}
  & \multicolumn{2}{c}{CA-2000}
  & \multicolumn{2}{c}{SC-2000}  \\
  \cmidrule(lr){2-3} \cmidrule(lr){4-5} \cmidrule(lr){6-7} \cmidrule(lr){8-9} \cmidrule(lr){10-11} \cmidrule(lr){12-13} \cmidrule(lr){14-15} \cmidrule(lr){16-17}
  Methods & Obj. $\downarrow$ & Gap $\downarrow$ & Obj. $\downarrow$ & Gap $\downarrow$ & Obj. $\downarrow$ & Gap $\downarrow$ & Obj. $\downarrow$ & Gap $\downarrow$ & Obj. $\downarrow$ & Gap $\downarrow$ & Obj. $\downarrow$ & Gap $\downarrow$  & Obj. $\downarrow$ & Gap $\downarrow$ & Obj. $\downarrow$ & Gap $\downarrow$ \\
  \midrule
  Gurobi    & 444.4 & -
            & -684.5 & -
            & -65017 & -
            & 293.0 & -
            & 681.9 & -
            & -453.2 & -
            & -63588 & -
            & 16.8 & -\\
  \midrule
  IL-LNS + SCIP & 443.5 & -0.42\scriptsize$\pm$0.74
                & \textbf{-684.3} & \textbf{0.04\scriptsize$\pm$0.08}
                & \textbf{-64745} & \textbf{0.42\scriptsize$\pm$0.74}
                & {292.6} & -0.14\scriptsize$\pm$0.69
                & 683.0 & 0.16\scriptsize$\pm$0.24
                & -458.0 & -1.07\scriptsize$\pm$0.97
                & {-63064} & 0.82\scriptsize$\pm$1.20
                & 198.0 & 1270.4\scriptsize$\pm$1963.4\\
  CL-LNS + SCIP & 443.5 & -0.20\scriptsize$\pm$0.30
                & -684.1 & 0.06\scriptsize$\pm$0.10
                & \underline{-64622} & \underline{0.60\scriptsize$\pm$1.05}
                & 294.1 & 0.39\scriptsize$\pm$0.87
                & 683.6 & 0.25\scriptsize$\pm$0.24
                & -457.3 & -0.92\scriptsize$\pm$1.93
                & \underline{-63082} & \underline{0.80\scriptsize$\pm$0.98}
                & 137.8 & 690.3\scriptsize$\pm$1348.4\\
  \midrule
  \textbf{MLBP}\\
  \textbf{\hspace{2pt} + SA} & \textbf{442.5} & \textbf{-0.43\scriptsize$\pm$0.39}
                    & -683.6 & 0.14\scriptsize$\pm$0.13
                    & -63835 & 1.82\scriptsize$\pm$0.57
                    & {291.6} & {-0.46\scriptsize$\pm$0.59}
                    & 683.7 & 0.26\scriptsize$\pm$0.17
                    & -454.9 & -0.37\scriptsize$\pm$0.98
                    & -62695 & 1.40\scriptsize$\pm$0.59
                    & 9.4 & -36.27\scriptsize$\pm$24.78\\
  \textbf{\hspace{2pt} + SA + {\small Reheat}} & \textbf{442.5} & \textbf{-0.43\scriptsize$\pm$0.39}
                    & {-684.0} & {0.08\scriptsize$\pm$0.10}
                    & -63780 & 1.88\scriptsize$\pm$2.17
                    & \textbf{291.5} & \textbf{-0.49\scriptsize$\pm$0.62}
                    & 683.4 & 0.22\scriptsize$\pm$0.16
                    & -459.0 & -1.29\scriptsize$\pm$0.83
                    & -62665 & 1.45\scriptsize$\pm$0.59
                    & 9.6 & -35.01\scriptsize$\pm$25.14\\
  \textbf{\hspace{2pt} + SA + $\tau$-PT} & {442.6} & {-0.39\scriptsize$\pm$0.40}
                                & {-683.8} & {0.10\scriptsize$\pm$0.10}
                                & {-63927} & {1.65\scriptsize$\pm$2.21}
                                & \underline{291.6} & \underline{-0.47\scriptsize$\pm$0.60}
                                & \textbf{681.7} & -0.04\scriptsize$\pm$0.13
                                & \underline{-459.1} & \underline{-1.30\scriptsize$\pm$0.83}
                                & \textbf{-63102} & \textbf{0.76\scriptsize$\pm$0.58}
                                & \textbf{9.3} & \textbf{-36.77\scriptsize$\pm$23.93}\\
  \textbf{\hspace{2pt} + SA + $\lambda$-PT} & {442.5} & {-0.41\scriptsize$\pm$0.40}
                                    & \underline{-684.2} & \underline{0.05\scriptsize$\pm$0.08}
                                    & {-63887} & {1.71\scriptsize$\pm$2.17}
                                    & {291.7} & {-0.42\scriptsize$\pm$0.71}
                                    & \underline{682.1} & \underline{0.03\scriptsize$\pm$0.16}
                                    & \textbf{-460.8} & \textbf{-1.69\scriptsize$\pm$0.87}
                                    & -62723 & 1.36\scriptsize$\pm$0.70
                                    & \textbf{9.3} & \textbf{-36.77\scriptsize$\pm$23.93}\\
  \bottomrule
\end{tabular}
}
\end{table*}

\begin{table*}[t]
    \centering
    \caption{Real-World MIPLIB instance results (objective values, the lower the better) within a 1000-second time budget. The best results among all methods, except Gurobi, are highlighted in \textbf{bold}.}
    \label{tab:realworld-results}
    \resizebox{0.7\textwidth}{!}{
    \begin{tabular}{lccccccc}
        \toprule
        \textbf{Instance} & \textbf{Gurobi} & \textbf{SCIP} & \textbf{IL-LNS} 
                          & \textbf{CL-LNS} & \textbf{SA} 
                          & \textbf{$\tau$-PT+SA} & \textbf{$\lambda$-PT+SA} \\
        \midrule
        cdc7-4-3-2     & -260.0 & -230.0 & $\times$ & $\times$ & \textbf{-244.0}& -241.0 & -243.0 \\
        cvs16r128-89 &-97.0&-95.0&\textbf{-97.0}&-96.0&-87.0&-88.0&-89.0\\
        d20200         & 12262.0 & \textbf{12289.0} &13318.0 & 13484.0 & 16062.0 & 13807.0 & 13612.0 \\
        eil33-2 &934.0&\textbf{934.0}&987.7&\textbf{934.0}&1626.9&1254.1&1344.3\\
        ex1010-pi      & 239.0 & {252.0} & \textbf{249.0} & 273.0 & 304.0 & 285.0 & 263.0 \\
        fast0507 &174.0&\textbf{174.0}&223.0&223.0&213.0&210.0&{200.0}\\
        queens-30      & -39.0 & -38.0 & \textbf{-39.0} &\textbf{ -39.0} & \textbf{-39.0} & \textbf{-39.0} & \textbf{-39.0} \\
        ramos3         & 238.0 & 242.0 & 231.0 & 219.0 & 225.0 & {224.0} & \textbf{217.0} \\
        scpj4scip      & 132.0 & 143.0 & 2942.0 & 2942.0 & \textbf{103.0} & 104.0 & 104.0 \\
        sorrell3       & -16.0 & -15.0 &  \textbf{-16.0} & \textbf{-16.0}& -15.0 & -15.0 & \textbf{-16.0} \\
        v150d30-2hopcds & 41.0 & \textbf{41.0} & \textbf{41.0} & \textbf{41.0} & \textbf{41.0} & \textbf{41.0} & \textbf{41.0} \\
        \bottomrule
    \end{tabular}
    }
    \vspace{-10pt}
\end{table*}

\subsection{Out-of-Distribution performance}
\label{sec:OOD}
\textbf{Research question}: \textit{Does the proposed sampling framework exhibit robust generalization to out-of-distribution instances compared to learning-based methods?}


\paragraph{Results.} We construct OOD test sets by shifting the underlying generation parameters 
of each problem class (Appendix~\ref{app:instance-generation}). As shown in the left half of 
\Cref{tab:ood_results1}, learning-based approaches achieve a distinct 
advantage on CA in the in-distribution setting, where the number of 
constraints is relatively small.
Since they rely on training data drawn from the training distribution, we expect them to degrade substantially under distribution shifts. Our sampling framework, by contrast, requires no training data; hyperparameter tuning is required under distribution shifts.

The right half of \Cref{tab:ood_results1} shows that learning-based methods suffer severe degradation on OOD instances whereas our sampling methods maintain competitive performance across all OOD tasks. 
By operating entirely without the need for training or label dependency, our sampling-based approach shows robust performance regardless of problem shifts.

\paragraph{Results on real-world instances.}
To validate practical applicability, we evaluate our 
framework on a heterogeneous set of MIPLIB 2017 \citep{gleixner2021miplib} instances 
with fewer than 10,000 constraints. All methods are given a 1000s time budget per instance. Our sampling methods use lightly-tuned hyperparameters ($\tau, \lambda$), chosen based on the scales of the objective and constraints, while the learning-based baselines (IL-LNS, CL-LNS) report the best performance among the pretrained models (each trained with one of MVC-1000, MIS-1500, CA-2000, or SC-2000). We mark an instance with `$\times$' if an algorithm cannot generate a feasible solution within the time budget. Full configurations of our methods are provided in Appendix~\ref{app:validation-tuning}.

As shown in \Cref{tab:realworld-results}, our framework attains performance competitive with classical solvers across this heterogeneous benchmark, despite using neither problem-specific heuristics nor task-specific training. In contrast, the learning-based baselines can fail significantly outside their training distribution (\textit{e.g.}, scpj4scip and fast0507), even though they utilize SCIP as a repair operator. These results support our central thesis: sampling-based optimization offers a robust, distribution-agnostic alternative to learning-based methods, and a useful complement to classical solvers on diverse real-world problems. However, in some instances, sampling can largely underperform compared to SCIP (\textit{e.g.}, eil33-2), and we believe this is because of either 1) less optimized implementation or 2) suboptimal hyperparameters. Combining sampling with heuristic solvers such as SCIP might address this weakness, which we leave as future work.

\subsection{Additional analysis}
\label{sec:additional analysis}

\begin{wraptable}[15]{r}{0.45\linewidth}
\vspace{-35pt}
\centering
\caption{Performance comparison of MLBP and baseline samplers under linear and squared penalty formulations.}
\label{tab:samplers}
\resizebox{\linewidth}{!}{
\begin{tabular}{ccccc}
  \toprule
  \addlinespace
  \multicolumn{5}{c}{(a) $\mathcal{E}(x;\lambda) = c^{\top}x + \lambda \sum_{k=1}^{m}\max(0, A_{k,\cdot} x - b_k)$} \\
  \addlinespace
  \midrule
  Proposal & MVC-2000 & MIS-3000 & CA-4000 & SC-4000 \\
  \midrule
  RWM    & Infeasible & -1216.25           & Infeasible          & 271.56           \\
  GWG    & Infeasible & Infeasible         & Infeasible          & 373.27           \\
  PAFS-3 & Infeasible & Infeasible         & Infeasible          & 339.29           \\
  PAFS-5 & Infeasible & Infeasible         & Infeasible          & 344.59           \\
  LBP  & 886.82     & -1367.07           & -124031.04          & \textbf{170.53}  \\
  MLBP-3  & 884.66     & -1369.10           & -126118.07          & 170.58           \\
  MLBP-5  & \textbf{884.53} & \textbf{-1369.14} & \textbf{-126657.27} & 172.08        \\
  \midrule
  \addlinespace
  \multicolumn{5}{c}{(b) $ \mathcal{E}(x;\lambda) = c^{\top}x + \lambda \sum_{k=1}^{m}\max(0, A_{k,\cdot} x - b_k)^2$} \\
  \addlinespace
  \midrule
  Proposal & MVC-2000 & MIS-3000 & CA-4000 & SC-4000 \\
  \midrule
  RWM     & Infeasible      & -1216.21            & Infeasible         & 271.56           \\
  GWG    & 899.70          & -1363.03           & Infeasible         & 171.75           \\
  PAFS-3 & Infeasible      & -1315.02           & Infeasible         & 186.49           \\
  PAFS-5 & Infeasible      & -1022.02           & Infeasible         & 227.62           \\
  LBP    & 886.36          & -1367.04           & -124017.53         & \textbf{170.53}  \\
  MLBP-3  & 884.72          & \textbf{-1369.22}  & -126085.10          & 170.66           \\
  MLBP-5  & \textbf{884.49} & -1369.15           & \textbf{-126636.02} & 172.08           \\
  \bottomrule
  \end{tabular}
}
\end{wraptable}

In this section, we evaluate our approach on double-sized instances. First, we conduct ablation studies to the proposal distribution. Second, we ablate the simulated annealing schedule to isolate its contribution within our framework. We also include the sensitivity analysis of the swap interval, the number of parallel chains in Appendix~\ref{app:additional}.

\paragraph{(M)LBP vs. Gradient-based proposals.} We argue that (M)LBP (\Cref{sec:proposal}) is more effective than gradient-based samplers for ILP, owing to its linear structure. To demonstrate this empirically, we compare (M)LBP, against existing discrete samplers, including Random Walk Metropolis (RWM), Gibbs-with-Gradients \citep[GWG;][]{grathwohl2021oops}, and the Path Auxiliary Fast Sampler \citep[PAFS;][]{sun2021path}. Additionally, we evaluate PAFS and MLBP for path lengths 3 and 5 (note that PAFS and MLBP with path length 1 are equivalent to GWG and LBP, respectively).
All samplers employ a Simulated Annealing (SA) schedule and are evaluated using the identical hyperparameters detailed in Appendix~\ref{app:validation-tuning}.

As illustrated in \Cref{tab:samplers}, (M)LBP finds feasible solutions across all four tasks under both penalty formulations, whereas gradient-based methods struggle substantially. Under the linear penalty ((a) in the table), GWG and PAFS remain entirely infeasible across all tasks. This is consistent with the piecewise-linear nature of the ILP energy: its gradient is piecewise constant and provides little local guidance, rendering first-order approximations of LBP essentially uninformative. RWM is feasible only on MIS and SC and infeasible on CA and MVC due to low penalty value and dense constraints respectively. Gradient-based methods partially recover under the squared penalty, making GWG feasible on MVC and MIS, but remain far below MLBP in solution quality. This confirms that the fundamental limitation lies in gradient approximation.

Regarding path length, increasing it from LBP (MLBP-1) to MLBP-5 consistently improves the objective on MVC-2000, MIS-3000, and CA-4000, while slightly degrading performance on SC-4000. This degradation arises because, in SC, flipping a single variable simultaneously affects multiple constraints, sharply reducing the MH acceptance rate as $L$ increases (see Appendix~\ref{app:acceptance_rate}). These results establish path length as a tunable knob for trading off local accuracy against global coverage, while (M)LBP remains feasibility-robust regardless of this choice.

\begin{table}[h]
    \vspace{-5pt}
  \centering
\caption{Ablation of the simulated annealing schedule under $\tau$-PT and 
$\lambda$-PT across four ILP benchmarks. All methods use the same MLBP 
proposal and runtime budget (200 seconds).}
  \label{tab:ablation_SA}
  \resizebox{0.9\textwidth}{!}{
  \begin{tabular}{l cc cc cc cc}
    \toprule

    & \multicolumn{2}{c}{MVC-2000}
    & \multicolumn{2}{c}{MIS-3000}
    & \multicolumn{2}{c}{CA-4000}
    & \multicolumn{2}{c}{SC-4000}  \\
    \cmidrule(lr){2-3} \cmidrule(lr){4-5} \cmidrule(lr){6-7} \cmidrule(lr){8-9} 
    Methods & Obj. $\downarrow$ & Gap $\downarrow$ & Obj. $\downarrow$ & Gap $\downarrow$ & Obj. $\downarrow$ & Gap $\downarrow$& Obj. $\downarrow$& Gap $\downarrow$\\
    \midrule
    Gurobi  & 897.3 & - & -1372.3 & - & -128231 & - & 170.9 & -\\
    \midrule
    \textbf{MLBP}\\
    \textbf{\hspace{1pt} + SA + $\tau$-PT}
                                  & \textbf{883.6} &\textbf{-1.52\scriptsize$\pm$0.42}
                                  & -1369.0 & 0.25\scriptsize$\pm$0.11
                                  & {-126212} & {1.57\scriptsize$\pm$0.81}
                                  & 172.0 & 0.68\scriptsize$\pm$1.06 \\
    \textbf{\hspace{1pt} + SA + $\lambda$-PT} 
                                      & 885.3  & -1.34\scriptsize$\pm$0.43
                                      & \textbf{-1370.5} &  \textbf{0.13\scriptsize$\pm$0.08}
                                      & \textbf{-126283} &   \textbf{1.51\scriptsize$\pm$0.85}
                                      & \textbf{170.3}  & \textbf{-0.36\scriptsize$\pm$0.69} \\
    \midrule
    \textbf{\hspace{1pt} + $\tau$-PT}
                                  & 883.6&-1.53\scriptsize$\pm$0.41
                                  &-1365.9 &0.47\scriptsize$\pm$0.13
                                  &Infeasible &-
                                  & 183.0 &7.18\scriptsize$\pm$2.31\\
    \textbf{\hspace{1pt} + $\lambda$-PT} 
                                      & 886.2&-1.23\scriptsize$\pm$0.45
                                      & -1367.2&0.38\scriptsize$\pm$0.09
                                      &-122335 &4.59\scriptsize$\pm$0.8
                                      & 180.1&5.47\scriptsize$\pm$2.38\\
    \bottomrule
  \end{tabular}
  }
  \vspace{-5pt}
\end{table}

\paragraph{Ablation of the annealing schedule.}
Throughout all experiments, three variants of our framework (SA, $\tau$ -PT + SA, and $\lambda$ -PT + SA) employ simulated annealing. We ablate the annealing component to verify whether SA is essential. As shown in \Cref{tab:ablation_SA}, removing SA leads to substantial degradation across all four benchmarks. 
These results confirm that SA is an indispensable component of our framework which first enables broad exploration and then concentrates the chains on high-quality feasible regions. 

\section{Conclusion}\label{sec:conclusion}
\paragraph{Contribution.}
We presented a training-free, solver-free framework that reformulates integer linear programming (ILP) as a discrete sampling problem. By exploiting the linear structure of ILP, we derive an efficient Multi-step Locally-Balanced Proposal (MLBP). Our sampling-based approach is both efficient and more robust than learning-based methods 
on out-of-distribution instances. This proposal outperforms gradient-based samplers across all benchmarks. To improve escape from local optima in the multimodal and constraint-fragmented energy landscapes of ILP, we employ two Parallel Tempering strategies, namely temperature-based ($\tau$-PT) and penalty-based ($\lambda$-PT). Experiments on four standard ILP benchmarks show that our method is competitive with Gurobi, the most advanced commercial ILP solver, consistently outperforms SCIP, and maintains strong performance on out-of-distribution instances where learning-based methods degrade significantly.

\paragraph{Limitation.}

Despite these results, several limitations remain. First, the convergence and mixing properties of Parallel Tempering in discrete, high-dimensional spaces are not yet well understood theoretically, leaving a gap between empirical effectiveness and formal guarantees. Second, the key hyperparameters, $\tau$ and $\lambda$, are selected via grid search on a small validation set guided by the scales of the objective and constraints, and minor adjustments may be needed when applying the framework to new problem classes. Finally, the framework relies on GPU-accelerated parallel computation, so performance may vary across hardware environments.


\bibliographystyle{unsrtnat}
\bibliography{reference}
\clearpage
\appendix

\section{Task details}
\label{app:task-details}
Following \citet{huang2023searching}, we evaluate our approach across four NP-hard binary Integer Programming Problems: Minimum Vertex Cover (MVC), Maximum Independent Set (MIS), Set Covering (SC), and Combinatorial Auction (CA). For each problem, we generate 20 and 100 instances as validation and testing sets, respectively. To further assess scalability, we generate double-sized test sets for each problem class. We denote these large-scale instances by appending their variable counts to the problem acronym (e.g., MVC-2000, MIS-3000, CA-4000, and SC-4000); comprehensive specifications for these scales are provided in \Cref{tab:instance-scale}. Detailed formulations and standard Lagrangian relaxations for each task are presented below.

\begin{table}[H]
\centering
\caption{Average number of variables and constraints across benchmark instances.}
\label{tab:instance-scale}
\vspace{8pt}
\small
\begin{tabular}{lcccc}
\toprule
Name & MVC-1000 & MIS-1500 & CA-2000 & SC-2000 \\
\midrule
Variables   & 1000  & 1500 & 2000 & 2000 \\
Constraints & 65100 & 6349 & 1371 & 5000 \\
\midrule
Name& MVC-2000 & MIS-3000 & CA-4000 & SC-4000  \\
\midrule
Variables& 2000 & 3000 & 4000  & 4000  \\
Constraints& 132000 & 12750 & 2675 & 5000   \\
\bottomrule
\end{tabular}
\end{table}
\subsection{Formulations}
\textbf{Minimum Vertex Cover (MVC)} \citep{huang2024contrastive}\\
Given an undirected graph $G = (V, E)$, the problem is to select the smallest subset of nodes $V' \subseteq V$ such that at least one endpoint of every edge in $G$ is selected:
\begin{align}
\min \quad & \sum_{v \in V} x_v \\ \nonumber
\text{s.t.} \quad
& x_u + x_v \geq 1, \quad \forall (u, v) \in E, \\\nonumber
& x_v \in \{0,1\}, \quad \forall v \in V.\nonumber
\end{align}
The standard Lagrangian formulation is
\begin{equation}
f(x):= \sum_{v \in V} x_v+\sum_{(u,v)\in E}\lambda_{uv}(1-x_u-x_v)
\end{equation}
where $\lambda_{uv}$ denotes the Lagrangian multiplier.

\textbf{Maximum Independent Set (MIS)} \citep{huang2024contrastive}\\
Given an undirected graph $G = (V, E)$, the problem is to select the largest subset of nodes $V' \subseteq V$ such that no two nodes in the subset are connected by an edge in $G$:
\begin{align}
\min \quad & -\sum_{v \in V} x_v \\ \nonumber
\text{s.t.} \quad
& x_u + x_v \leq 1, \quad \forall (u, v) \in E, \\\nonumber
& x_v \in \{0,1\}, \quad \forall v \in V.\nonumber
\end{align}

The standard Lagrangian formulation is
\begin{equation}
f(x):= -\sum_{v \in V} x_v+\sum_{(u,v)\in E}\lambda_{uv}(x_u+x_v-1)
\end{equation}
where $\lambda_{uv}$ denotes the Lagrangian multiplier.




\textbf{Set Covering (SC)} \citep{huang2026general}\\
In the SC problem, given $m$ elements and $n$ sets of elements, we aim to cover all elements using as few sets as possible. The union of $n$ sets forms a collection $S$ covering m elements, each of which belongs to at least one set. The SC problem can be formulated as follows:
\begin{align}
\min \quad & \sum_{i=1}^n x_i, \\ \nonumber
\text{s.t.} \quad
& \sum_{i:j \in s_i} x_i \ge 1, \quad j=1, \dots,m \\ \nonumber
& x_i \in \{0,1\}, \quad  i=1, \dots,n. 
\end{align}
where $x_i$ denotes whether the $i$-th set $s_i$ is selected. The set of constraints~(2) ensures that each item is covered by at least one set, and the constraints~(3) impose integrality on the variables.

The standard Lagrangian formulation is
\begin{equation}
f(x):=\sum_{i=1}^n x_i + \sum_{j=1}^{m} \lambda_j \left( 1- \sum_{i:j \in s_i} x_i \right)
\end{equation}
where $\lambda_j$ denotes the Lagrangian multiplier for constraint $j$.

\textbf{Combinatorial Auction (CA)} \citep{huang2024contrastive}\\
Given $n$ bids $\{(B_i, p_i) : i \in [n]\}$ for $m$ items, where $B_i$ is a subset of items and $p_i$ is the bidding price for $B_i$, we want to allocate items to bids such that the total revenue is maximized:
\begin{align}
\min \quad & -\sum_{i \in [n]} p_i x_i \\
\text{s.t.} \quad
& \sum_{i : j \in B_i} x_i \le 1, \quad \forall j \in [m], \\
& x_i \in \{0,1\}, \quad \forall i \in [n].
\end{align}

The standard Lagrangian formulation is
\begin{equation}
f(x):=-\sum_{i \in [n]} p_i x_i+\sum_{j \in [m]}\lambda_j\left(\sum_{i : j \in B_i} x_i - 1\right)
\end{equation}
where $\lambda_j$ denotes the Lagrangian multiplier.

\subsection{Instance generation details}
\label{app:instance-generation}
All benchmark instances are generated using the Ecole library \footnote{\url{https://doc.ecole.ai/py/en/stable/reference/instances.html}} based on established configurations. Specifically, Minimum Vertex Cover (MVC) instances are created via the Barabási-Albert random graph model \citep{albert2002statistical} with 1,000 nodes and an average degree of 70, following \citet{song2020general}. Similarly, Maximum Independent Set (MIS) instances follow \citet{song2020general}, utilizing the Erdős-Rényi model \citep{erdos1961evolution} to produce graphs with 1,500 nodes and an average degree of 5. For Combinatorial Auctions (CA), we adopt the arbitrary relations distribution introduced by \citet{leyton2000towards} to generate instances with 1,000 items and 2,000 bids. Set Cover (SC) instances, following \citet{wu2021learning}, contain 2,000 variables and 5,000 constraints. To further evaluate scalability, we construct an additional test set of 100 large-scale instances for each problem class by doubling the variable count. The average number of variables and constraints for all generated sets is summarized in \Cref{tab:instance-scale}.

\textbf{Out-of-Distribution instances.}
To examine generalizability, we construct an Out-of-Distribution (OOD) test set of $20$ instances per problem class by modifying their inherent generation parameters. Specifically, for MVC, we significantly reduce the average degree from $70$ to $5$. For MIS, we shift the underlying graph structure to the Barabási-Albert model. For SC, we increase the density parameter (the density of the constraint matrix) from $0.05$ to $0.5$, and for CA, we increase the max-n-sub-bids parameter (the maximum number of substitutable bids per bidder) from $5$ to $15$.

\clearpage

\section{Non-reversible parallel tempering}
\label{app:non-reversible}
Sampling in discrete domains is more prone to being trapped in local optima than in continuous domains \citep{li2025reheated, feng2025regularized}. Parallel Tempering (PT) is a standard remedy, and the efficiency of PT relies on the round-trip rate \citep{syed2022non}. In this paper, we adopt the non-reversible parallel tempering scheme proposed by \citet{syed2022non}, which facilitates more frequent swaps between adjacent chains than reversible PT.

\textbf{Communication kernels.}
Before introducing the communication scheme, we first define its fundamental building block, namely a \emph{swap}. A swap is a Metropolis–Hastings transition with a deterministic proposal that exchanges the states of two adjacent chains.

For $n = 1, \ldots, B$, the swap proposal $x^{(n,n+1)}$ is defined as
\begin{equation}
x^{(n,n+1)} = (x^1, \ldots, x^{n-1}, x^{n+1}, x^{n}, x^{n+2}, \ldots, x^B),
\end{equation}
which transposes the $n$-th and $(n+1)$-th components of $\mathbf{x}$. The associated Metropolis–Hastings kernel $K^{(n,n+1)}(\mathbf{x}, \mathrm{d}\mathbf{x}')$ is given by
\begin{equation}
K^{(n,n+1)}(\mathbf{x}, \mathrm{d}\mathbf{x}')
= \bigl(1 - \alpha_{\text{swap}}(x^n, x^{n+1})\bigr)\,\delta_{\mathbf{x}}(\mathrm{d}\mathbf{x}') 
+ \alpha_{\text{swap}}(x^n, x^{n+1})\,\delta_{\mathbf{x}^{(n,n+1)}}(\mathrm{d}\mathbf{x}'),
\end{equation}
where $\alpha_{\text{swap}}$ is the swap acceptance probability defined in 
\Cref{temperature_swap} (or \Cref{penalty_swap} for $\lambda$-PT).

\textbf{Non-reversible parallel tempering.}
A critical design choice in Parallel Tempering is the selection of the $(n, n+1)$ pairs for communication. While traditional reversible PT selects these pairs at random, \citet{syed2022non} introduce non-reversibility through a deterministic alternation of "even" and "odd" kernels:
\begin{equation}
K^{\text{even}}:=\prod_{n \, \text{even}}K^{(n,n+1)}, \quad K^{\text{odd}}:=\prod_{n \, \text{odd}}K^{(n,n+1)},
\end{equation}
The algorithm then follows a Deterministic-Even-Odd (DEO) kernel, $K_t^{\text{DEO}}$, which iterates between "even" and "odd" kernels deterministically based on the time step $t$:
\begin{equation}
\label{kernel_DEO}
K_t^{\mathrm{DEO}} :=
\begin{cases}
K^{\mathrm{even}} & \text{if } t \text{ is even}, \\
K^{\mathrm{odd}}  & \text{if } t \text{ is odd}.
\end{cases}
\end{equation}
This deterministic switching between even and odd kernels provably attains a higher round-trip rate and improved mixing efficiency than reversible PT \citep{syed2022non}.
\clearpage
\section{Algorithms}
\label{app:algorithms}
\begin{algorithm}
\caption{MLBP with Parallel Tempering}
\label{alg:lbp}
\begin{algorithmic}

\STATE \textbf{Input:} $A \in \mathbb{R}^{m \times n}$, $c \in \mathbb{R}^n$, $b \in \mathbb{R}^{m}$, $\Lambda$, $\bar{\tau}$
\STATE
\STATE Initialize $\{x_{(0)}^i\}_{i=1}^{B}$ by $\mathrm{Bernoulli}(0.5)$;
\STATE $s \leftarrow 0$;\quad $x^{\star} \leftarrow x^1$
\FOR{$t=0, \dots, T$}
    \FOR{each chain $i \in \{1,\dots,B\}$ \textbf{in parallel}}
        \STATE $\triangleright$ \textbf{Sample candidate $y^i$}
        \STATE $d_n(x^i) \leftarrow \dfrac{\mathcal{E}(x^i) - \mathcal{E}((x^i)^{-n})}{\tau^i} \quad \forall n$
        \STATE $w_n(x^i) \leftarrow {g}\!\left(\exp(d_n(x^i))\right) \quad \forall n$
        \STATE $J \sim \mathrm{Categorical}\!\left(\dfrac{w_n(x^i)}{\sum_n w_n(x^i)}\right)$ 
        \STATE $y^i \leftarrow x^i$;\quad $y^i_j \leftarrow 1 - x^i_j \quad \forall j \in J$ \COMMENT{Flip bits}
        \STATE
        \STATE $\triangleright$ \textbf{MH acceptance}
        \STATE $w_n(y^i) \leftarrow  {g}\!\left(\exp\!\left(\dfrac{\mathcal{E}(y^i) - \mathcal{E}((y^i)^{-n})}{\tau^i}\right)\right)$ $\forall n$
        \STATE $\bar{w}_n(z) \leftarrow \dfrac{w_n(z)}{\sum_n w_n(z)}$ \, for $z \in \{x^i, y^i\}$, $\forall n$
        \STATE $R \leftarrow \exp \left( \sum_{j \in J} d_j(x^i) \right) \cdot \prod_{j \in J} \frac{\bar{w}_j(y^i)}{\bar{w}_j(x^i)}$
        \STATE $\alpha(y^i, x^i) \leftarrow \min(1, R)$
        \STATE $x^i \leftarrow y^i$ with probability $\alpha(y^i, x^i)$
    \ENDFOR
    \STATE
    \STATE $\triangleright$ \textbf{Parallel Tempering}
    \IF{$t \bmod I = 0$}
        \IF{$s$ is even}
            \STATE $\mathcal{S} \leftarrow \{(1,2),(3,4),\dots\}$
        \ELSE
            \STATE $\mathcal{S} \leftarrow \{(2,3),(4,5),\dots\}$
        \ENDIF
        \FOR{each pair $(i,i') \in \mathcal{S}$ \textbf{in parallel}}
            \STATE $x^i \leftrightarrow x^{i'}$ with probability $\alpha_{swap}(x^i,x^{i'})$
        \ENDFOR
        \STATE $s \leftarrow s + 1$
    \ENDIF
    \STATE $x^{best} \leftarrow \arg\min_{i: A x^i \leq b}\, \mathcal{E}(x^i)$  
       \COMMENT{Best feasible solution among chains}
    \IF{$\mathcal{E}(x^{best}) < \mathcal{E}(x^{\star})$}
        \STATE $x^{\star} \leftarrow x^{best}$
\ENDIF
\ENDFOR
\STATE \textbf{Output:} Best feasible Solution $x^\star$

\end{algorithmic}
\end{algorithm}

\clearpage

\section{Experimental details}
\subsection{Hyperparameter tuning for temperature and penalty parameter}
\label{app:validation-tuning}
MCMC algorithms for constrained optimization depend heavily on the temperature ($\tau$) and penalty parameter ($\lambda$), which we tune using 20 validation instances. To reduce the tuning burden of the additional hyperparameter in our $\tau$-PT+SA and $\lambda$-PT+SA algorithms, we leverage results from the SA schedule. Specifically, we first tune $\tau$ and $\lambda$ for {SA} in \Cref{tab:validation-tuning-exp-decay}. Then, $\tau$-PT+SA only tunes the temperature ladder $\bar{\tau}=(\tau_{min},\tau_{max})$ while fixing the penalty, and $\lambda$-PT+SA only tunes the penalty ladder $\Lambda$ while fixing the temperature. We apply rule-based configurations: $\bar{\tau}=(\tau, 2\tau)$ for $\tau$-PT+SA and $\Lambda = (\lambda/2, \lambda)$ for $\lambda$-PT+SA, as shown in 
\Cref{tab:validation-tuning-pt}. Across all experiments, we employ a geometrically spaced temperature ladder with 15 chains and fix the swap interval $I=200$ and decay rate $\gamma$ such that $\gamma^{100000}=0.5$. We narrow the grid search space of $\tau$ and $\lambda$ by aligning their scales 
with the objective function, testing 3–4 values per parameter. The best hyperparameters identified on small-scale instances are directly transferred to large-scale instances without re-tuning. We report the final hyperparameters in \Cref{tab:hyperparameter-settings}.

\begin{table}[htbp]
    \centering
    \caption{Validation tuning results for temperature ($\tau$) and penalty parameter ($\lambda$) in SA schedules.}
    \label{tab:validation-tuning-exp-decay}
    
    \resizebox{\textwidth}{!}{
        \begin{tabular}{@{}lccccccccccccccc@{}}
            \toprule
            \multicolumn{4}{c}{{MVC-1000}} & \multicolumn{4}{c}{{MIS-1500}} & \multicolumn{4}{c}{{CA-2000}} & \multicolumn{4}{c}{{SC-2000}} \\ 
            \cmidrule(r){1-4} \cmidrule(lr){5-8} \cmidrule(lr){9-12} \cmidrule(l){13-16}
            
            $\tau \setminus \lambda$ & 1.0&2.0 & 5.0&$\tau \setminus \lambda$& 1.0& 2.0&5.0 &$\tau \setminus \lambda$ &300.0&400.0 & 500.0&$\tau \setminus \lambda$ &1.0&2.0 & 5.0\\ \midrule
            
            0.1 &444.96&453.55&454.35&0.1&Infeasible&-613.40&-613.60&10&Infeasible&-51069.58&-51276.05&0.1&932.45&843.60&611.34\\
            
            0.2 &\textbf{443.70}&444.32&444.68&0.2&Infeasible&\textbf{-683.75}&-682.15&20&-59091.29&-55044.77&-53226.79&0.2&944.45&846.65&552.95\\
            
            0.5 &Infeasible&450.97&450.49&0.5&Infeasible&-673.80&-675.20&50&\textbf{-64332.64}&-63381.85&-62369.05&0.5&996.45&892.60&298.15\\ 

            1.0 &Infeasible&464.69&458.87&1.0&Infeasible&Infeasible&606.75&100&-59436.90&-61394.33&-61247.26&1.0&1035.55&906.75&\textbf{284.65}\\ 

            \bottomrule
        \end{tabular}
    }
\end{table}

\begin{table}[htbp]
    \centering
    \caption{Validation tuning results for temperature and penalty ladders.}
    \label{tab:validation-tuning-pt}

    \begin{subtable}{0.49\textwidth}
        \centering
        \caption{$\tau$-PT+SA}
        \resizebox{\textwidth}{!}{
        \begin{tabular}{@{}lccccccc@{}}
            \toprule
            \multicolumn{2}{c}{MVC-1000} & \multicolumn{2}{c}{MIS-1500} & \multicolumn{2}{c}{CA-2000} & \multicolumn{2}{c}{SC-2000} \\ 
            \cmidrule(r){1-2} \cmidrule(lr){3-4} \cmidrule(lr){5-6} \cmidrule(l){7-8}
            $\bar{\tau} \setminus \lambda$ & 1.0 & $\bar{\tau} \setminus \lambda$  & 2.0  & $\bar{\tau} \setminus \lambda$ & 300.0 & $\bar{\tau} \setminus \lambda$ & 5.0 \\ 
            \midrule
            (0.1,0.2) & \textbf{443.70}  & (0.1,0.2)  &  -677.50 & (10.0,20.0)  &-55602.11 & (0.1,0.2) &595.80  \\
            (0.2,0.4) &443.82 & (0.2,0.4) & \textbf{-683.85} & (20.0,40.0)  &-62205.26& (0.2,0.4) & 399.75 \\
            (0.5,1.0) & Infeasible & (0.5,1.0)  & -668.15  & (50.0,100.0)  &  \textbf{-64406.34}& (0.5,1.0)& 286.50 \\ 
            (1.0,2.0) & Infeasible & (1.0,2.0) &   Infeasible & (100.0,200.0)   & Infeasible & (1.0,2.0) & \textbf{284.59}\\ 
            \bottomrule
        \end{tabular}}
    \end{subtable}
    \hfill
    \begin{subtable}{0.49\textwidth}
        \centering
        \caption{$\lambda$-PT+SA}
        \resizebox{\textwidth}{!}{
        \begin{tabular}{@{}lccccccc@{}}
            \toprule
            \multicolumn{2}{c}{MVC-1000} & \multicolumn{2}{c}{MIS-1500} & \multicolumn{2}{c}{CA-2000} & \multicolumn{2}{c}{SC-2000} \\ 
            \cmidrule(r){1-2} \cmidrule(lr){3-4} \cmidrule(lr){5-6} \cmidrule(l){7-8}
            $\Lambda \setminus \tau$ & 0.2 & $\Lambda \setminus \tau$  & 0.2  & $\Lambda \setminus \tau$ & 50.0 & $\Lambda \setminus \tau$ & 1.0 \\ 
            \midrule
            (0.5,1.0) &\textbf{443.71}& (0.5,1.0)  &Infeasible   & (150.0,300.0)  & Infeasible & (0.5,1.0) & 1035.55  \\
            (1.0,2.0) & 443.82& (1.0,2.0) &\textbf{-684.30}  & (200.0,400.0)  &\textbf{-64200.59}& (1.0,2.0) &977.35  \\
            (2.5,5.0) & 444.50 & (2.5,5.0)  &  -682.35& (250.0,500.0)  & -63944.81 & (2.5,5.0)&  \textbf{284.85} \\ 
            \bottomrule
        \end{tabular}}
    \end{subtable}
\end{table}

\begin{table}[ht]

\centering
\caption{Final hyperparameters for SA, $\tau$-PT+SA, and $\lambda$-PT+SA.}
\label{tab:hyperparameter-settings}
\resizebox{0.6\linewidth}{!}{
\begin{tabular}{c c c c c c}\toprule
& Schedule
&Temperature $\tau$
&Penalty parameter $\lambda$
\\\midrule
MVC      & SA   &0.2  &$1$\\
MIS      & SA   &0.2   &$2 $\\
SC       & SA    &1.0   &$5$\\
CA        & SA    &50.0   &$300$\\
\midrule
& Schedule
&Temperature ladder $\bar{\tau}$
&Penalty parameter $\lambda$
\\\midrule
MVC      & $\tau$-PT+SA   &(0.1,0.2)   &$1$\\
MIS      & $\tau$-PT+SA    &(0.2,0.4)   &$ 2$\\
SC       & $\tau$-PT+SA    &(1.0,2.0)   &$5$\\
CA       & $\tau$-PT+SA    &(50.0,100.0)   &$300$\\
\midrule
& Schedule
&Temperature $\tau$
&Penalty ladder $\Lambda$
\\\midrule
MVC      & $\lambda$-PT+SA   &0.2   &$(0.5,1) $\\
MIS      & $\lambda$-PT+SA   &0.2   &$(1,2) $\\
SC        & $\lambda$-PT+SA    &1.0   &$(2.5,5)$\\
CA       & $\lambda$-PT+SA    &50.0   &$(200,400)$\\
\bottomrule
\end{tabular}
}
\end{table}

\paragraph{Hyperparameters for MIPLIB 2017.}
For the real-world MIPLIB 2017 experiments, we follow the same procedure as before: only $\tau$ and $\lambda$ for SA are tuned, and the corresponding ladders for $\tau$-PT+SA and $\lambda$-PT+SA are derived directly from these values via the rule-based scheme. This keeps the tuning cost minimal; the final hyperparameters are listed in \Cref{tab:realworld-hyperparameter-settings}.

\begin{table}[ht]
\centering
\caption{Final hyperparameters for SA, $\tau$-PT+SA, and $\lambda$-PT+SA on MIPLIB 2017.}
\label{tab:realworld-hyperparameter-settings}
\resizebox{0.8\linewidth}{!}{
\begin{tabular}{l cc cc cc}
\toprule
& \multicolumn{2}{c}{SA} & \multicolumn{2}{c}{$\tau$-PT+SA} & \multicolumn{2}{c}{$\lambda$-PT+SA} \\
\cmidrule(lr){2-3} \cmidrule(lr){4-5} \cmidrule(lr){6-7}
Instance        & $\tau$ & $\lambda$ & $\bar{\tau}$ & $\lambda$ & $\tau$ & $\Lambda$ \\
\midrule
cdc7-4-3-2      & 0.5 & 2.0 &(0.5,1.0)  & 2.0 & 0.5 & (1.0,2.0)  \\
cvs16r128-89    &  0.5 &  5.0 & (0.5,1.0) & 5.0  & 0.5  & (2.5,5.0)  \\
d20200          &20.0  & 200.0 &(20.0,40.0)  &200.0  & 20.0 & (100.0,200.0) \\
eil33-2         &400.0  &1000.0  & (400.0,800.0) &1000.0  &400.0  &(500.0,1000.0) \\
ex1010-pi       &0.1  & 1.0 & (0.1,0.2) & 1.0 & 0.1 & (0.5,1.0) \\
fast0507        & 0.2  & 2.0  & (0.2,0.4) &2.0  & 0.2 & (1.0,2.0) \\
queens-30       & 0.5 &2.0& (0.5,1.0) &2.0  & 0.5 &(1.0,2.0)  \\
ramos3          & 0.5 &2.0& (0.5,1.0) &2.0  & 0.5 &(1.0,2.0)  \\
scpj4scip       &0.1  & 1.0 & (0.1,0.2) & 1.0 & 0.1 & (0.5,1.0)\\
sorrell3        & 0.2  & 2.0  & (0.2,0.4) &2.0  & 0.2 & (1.0,2.0)  \\
v150d30-2hopcds & 0.2 & 1.0  &(0.2,0.4)  &1.0  &0.2  &(0.5,1.0)  \\
\bottomrule
\end{tabular}
}
\end{table}

\subsection{Total MCMC time steps}
\label{app:MCMC-steps}
We conduct all experiments on a single NVIDIA L40S 1 GPU and AMD EPYC 9554 64-Core Processor CPU, with a runtime limit of 200 seconds per instance. \Cref{tab:total-time-steps} presents the total MCMC time steps completed within a fixed 200-second runtime for the SA baseline and our proposed $\tau$-PT+SA ($\lambda$-PT+SA) schedules. Across four tasks (MVC, MIS, SC, and CA), the results show that increasing the number of parallel chains reduces the total time steps due to the higher computational cost per iteration. Furthermore, the $\tau$-PT+SA ($\lambda$-PT+SA) schedules generally complete fewer time steps than the SA baseline, reflecting the increased computational demands of the parallel tempering process. Nonetheless, the gap between SA and PT-based schedules remains small (e.g., within 5–10\% on most tasks), confirming that the additional cost of parallel tempering is marginal.

\begin{table}[ht]
\centering
\caption{Total MCMC time steps executed within the 200-second runtime budget across varying numbers of parallel chains.}
\label{tab:total-time-steps}
\resizebox{0.7\linewidth}{!}{
\begin{tabular}{c c ccc}\toprule 
& Schedule
& \# chains$=5$ 
& \# chains$=15$
& \# chains$=25$
\\\midrule
MVC-2000      & SA &13,500& 10,000 &8,000\\
MIS-3000      & SA &65,000 &55,000& 47,500\\
SC-4000       & SA & 80,000&78,000&70,000\\
CA-4000        & SA &92,000 &88,000&87,500\\
\midrule
& Schedule
& \# chains$=5$ 
& \# chains$=15$
& \# chains$=25$
\\\midrule
MVC-2000      &$\tau$-PT+SA & 13,500&10,000&8,000\\
MIS-3000      &$\tau$-PT+SA & 60,000& 52,000&45,000\\
SC-4000       & $\tau$-PT+SA & 76,000&74,000&66,000\\
CA-4000       &$\tau$-PT+SA &87,500 &84,000&79,000\\
\bottomrule
\end{tabular}
}
\end{table}
\clearpage
\section{Baseline details}
\label{app:baselines-details}
In this section, we provide a detailed description of our baseline implementation.

\textbf{SCIP, Gurobi} \citep{achterberg2009scip, optimization2020gurobi}: We use the off-the-shelf implementations of SCIP and Gurobi as exact ILP solvers, with default parameter settings unless otherwise specified. For our experiments, Gurobi utilized 24 threads and achieves state-of-the-art results among the solvers.

\textbf{DINS, GINS, RINS, RENS} \citep{achterberg2007constraint}: These are classical primal heuristics integrated within the SCIP solver. We evaluate them as baseline Large Neighborhood Search (LNS) methods to provide a rigorous comparison between our approach and traditional, rule-based neighborhood selection strategies. For practical implementation, we utilize PySCIPOpt \footnote{\url{https://github.com/scipopt/PySCIPOpt}}, a Python interface for SCIP, which allows for the programmatic execution and parameter tuning of these specific LNS heuristics.

\textbf{IL-LNS}, \textbf{CL-LNS}  \citep{sonnerat2021learning, huang2023searching}: For learning-based approaches, we generate 1,024 small-scale instances for each problem and split these instances into training (896) and validation (128) sets. Training labels are generated using Local Branching (LB) \citep{fischetti2003local}. Following the data collection protocol of \cite{huang2023searching}, we adapt the primary hyperparameter—the neighborhood size $k_0$. While \cite{huang2023searching} utilized $k_0$ values of 50, 500, 200, and 50 for MVC, MIS, CA, and SC, respectively, we employ proportional values of 125, 100, and 25 for MIS, CA, and SC to account for the smaller variable counts in our problem set. LB is executed for 10 iterations with a one-hour runtime limit per iteration. Models are trained with a batch size of 32 over 30 epochs.

To evaluate the learning-based Large Neighborhood Search (LNS), we first obtain an initial solution by running SCIP \citep{achterberg2009scip} for 10 seconds. In each LNS iteration, a GNN-based destroy operator selects variables to be removed, and the resulting sub-ILP is then re-optimized using SCIP. For evaluation, we maintain proportional neighborhood sizes across tasks: 100 for MVC, 750 for MIS, 500 for CA, and 50 for SC. The detailed implementation follows the original codebase\footnote{\url{https://github.com/facebookresearch/CL-LNS}} to reproduce the results.
\clearpage
\clearpage
\section{Further experiments}
\label{app:additional}
\subsection{Swap interval $I$} In Parallel Tempering, chains at different temperatures periodically attempt to exchange their states, and the swap interval $I$ controls how frequently this exchange occurs: a state swap is proposed every $I$ MCMC steps. If $I$ is too small, swaps are attempted before individual chains explore their local neighborhood, resulting in highly correlated swap proposals that hinder efficient exploration. If $I$ is too large, the chains evolve nearly independently for long stretches, reducing the benefit of inter-chain communication and slowing global mixing.

In our experiments, as illustrated in \Cref{tab:swap-interval}, both $\tau$-PT+SA and $\lambda$-PT+SA maintain stable objective values across swap intervals of $I \in  \{50, 100, 200, 500\}$. The performance gap between the best and worst $I$ setting is within $0.03$ on MIS-3000 and $0.41$ on CA-4000 for $\tau$-PT+SA, and similarly narrow for $\lambda$-PT+SA. This robustness indicates that performance is primarily governed by the underlying exchange mechanism, rather than the precise cadence of swap proposals. Although marginal gains are occasionally achievable through tuning (e.g., $\tau$-PT+SA at $I=200$ achieves a slightly better gap on CA-4000 vs. $I=500$), the overall effect is negligible. We therefore fix $I=200$ as the default across all experiments.
\begin{table}[H]
\centering
\caption{Sensitivity analysis of the parallel tempering swap interval ($I$) across four benchmarks.}
\label{tab:swap-interval}
\resizebox{\linewidth}{!}{
\begin{tabular}{c cc cc cc cc}
  \toprule
  \addlinespace 
  \multicolumn{9}{c}{\large \textbf{$\tau$-PT+SA}} \\
  \addlinespace 
\midrule
  & \multicolumn{2}{c}{MVC-2000} & \multicolumn{2}{c}{MIS-3000} & \multicolumn{2}{c}{CA-4000} & \multicolumn{2}{c}{SC-4000}  \\
  \cmidrule(lr){2-3} \cmidrule(lr){4-5} \cmidrule(lr){6-7} \cmidrule(lr){8-9}
  $I$ & Obj. $\downarrow$& Gap $\downarrow$& Obj. $\downarrow$& Gap $\downarrow$& Obj. $\downarrow$& Gap $\downarrow$ & Obj. $\downarrow$& Gap $\downarrow$\\
  \midrule
  50            & 883.68 & -1.52\scriptsize$\pm$0.41 
                & -1369.08 & 0.24\scriptsize$\pm$0.10
                & -126099.32 & 1.66\scriptsize$\pm$0.89  
                & 171.94 & 0.65\scriptsize$\pm$0.99\\
  100           & 883.78 & -1.51\scriptsize$\pm$0.41 
                &  -1368.98   &  0.25\scriptsize$\pm$0.10
              & -126161.48 &1.61\scriptsize$\pm$0.87
              & 171.87 &0.60\scriptsize$\pm$0.95\\
  200 (default) & 883.64 & -1.52\scriptsize$\pm$0.42 
                & -1368.98&0.25\scriptsize$\pm$0.11 
                & -126212.32 & 1.57\scriptsize$\pm$0.81 
                & 172.00 & 0.68\scriptsize$\pm$1.06 \\
  500           & 883.50 & -1.54\scriptsize$\pm$0.41  
                & -1368.98 & 0.24\scriptsize$\pm$0.11
                & -126137.03 & 1.63\scriptsize$\pm$0.93
                & 172.27 &0.84\scriptsize$\pm$1.08 \\
  \midrule
  \addlinespace 
  \multicolumn{9}{c}{\large \textbf{$\lambda$-PT+SA}} \\
  \addlinespace 
\midrule
  & \multicolumn{2}{c}{MVC-2000} & \multicolumn{2}{c}{MIS-3000} & \multicolumn{2}{c}{CA-4000} & \multicolumn{2}{c}{SC-4000}  \\
  \cmidrule(lr){2-3} \cmidrule(lr){4-5} \cmidrule(lr){6-7} \cmidrule(lr){8-9}
  $I$ & Obj. $\downarrow$& Gap $\downarrow$& Obj. $\downarrow$& Gap $\downarrow$& Obj. $\downarrow$& Gap $\downarrow$ & Obj. $\downarrow$& Gap $\downarrow$\\
  \midrule
  50            & 885.21 &-1.35\scriptsize$\pm$0.40 
                & -1370.61 & 0.13\scriptsize$\pm$0.08
                & -126217.33 & 1.56\scriptsize$\pm$0.88
                & 170.31 &-0.32\scriptsize$\pm$0.67\\
  100           & 885.25 &-1.34\scriptsize$\pm$0.41 
                & -1370.68&  0.12\scriptsize$\pm$0.08
                & -126237.74  &1.55\scriptsize$\pm$0.74
                & 170.49 & -0.21\scriptsize$\pm$0.75\\
  200 (default) & 885.25& -1.34\scriptsize$\pm$0.43 
                &-1370.52& 0.13\scriptsize$\pm$0.08   
                & -126283.38 & 1.51\scriptsize$\pm$0.85   
                & 170.25 & -0.36\scriptsize$\pm$0.69 \\
  500           & 885.61 &-1.30\scriptsize$\pm$0.42 
                & -1370.45 & 0.14\scriptsize$\pm$0.09
                & -126087.47  & 1.67\scriptsize$\pm$0.82
                & 170.46 &-0.23\scriptsize$\pm$0.66 \\
  \bottomrule
\end{tabular}
}
\end{table}

\subsection{Number of chains}
 Regarding the number of chains (\Cref{tab:number-of-chains}), SA benefits monotonically from more chains since additional independent runs increase coverage of the objective landscape (e.g., MIS-3000 Gap improves from 0.31 at 5 chains to 0.20 at 25 chains). For PT methods, however, the number of chains determines the spacing of the tempering ladder, and this introduces a trade-off: too few chains yield wide ladder spacing that reduces swap acceptance, and in extreme cases may even prevent feasibility (CA-4000 becomes Infeasible at 5 chains for $\tau$-PT+SA). 

 Specifically, $\tau$-PT+SA on CA-4000 yields an infeasible result for a marginal 3\% of instances when using 5 chains. This phenomenon occurs because fewer chains widen the temperature gaps. Under the assumption of efficient swapping, exploring the infeasible space is necessary to achieve better performance, which is why a small penalty value of 300 was chosen as optimal. However, as the temperature gap between chains widens, swap acceptance drops significantly. Consequently, the search process in a few instances transitions predominantly within the infeasible space, failing to reach feasibility. Since this issue is extremely rare (3 instances among 100 instances), it can be easily circumvented by increasing the \# chains, adjusting the penalty parameter ($\lambda$), or modifying the temperature ladder. Furthermore, when excluding these few infeasible instances, the method still achieves a competitive average objective value of $-125411.65$.
 
 Conversely, too many chains produce densely spaced, nearly identical adjacent distributions where swaps become uninformative and per-chain step budgets are reduced (see Appendix~\ref{app:MCMC-steps}). The default of 15 chains provides a practical balance across all tasks and methods.
 \begin{table}[H]
\centering
\caption{ Sensitivity analysis of the number of parallel chains ($B$) across four benchmarks.}
\label{tab:number-of-chains}
\resizebox{\linewidth}{!}{
\begin{tabular}{c cc cc cc cc} 
  \toprule
   \addlinespace 
   \multicolumn{9}{c}{\large \textbf{SA}} \\
    \addlinespace 
     \midrule
   & \multicolumn{2}{c}{MVC-2000} & \multicolumn{2}{c}{MIS-3000} & \multicolumn{2}{c}{CA-4000} & \multicolumn{2}{c}{SC-4000} \\
  \cmidrule(lr){2-3} \cmidrule(lr){4-5} \cmidrule(lr){6-7} \cmidrule(lr){8-9}
   $B$ & Obj. $\downarrow$& Gap $\downarrow$& Obj. $\downarrow$& Gap $\downarrow$& Obj. $\downarrow$& Gap $\downarrow$& Obj. $\downarrow$& Gap $\downarrow$\\
   \midrule

   5    &886.44 & -1.21\scriptsize$\pm$0.43 
        &-1368.13&0.31\scriptsize$\pm$0.11
        &-125556.94&2.08\scriptsize$\pm$0.85
        &171.38 & 0.31\scriptsize$\pm$0.93 \\
   15 (default.)  & 884.66& -1.41\scriptsize$\pm$0.41
        &-1369.10&0.24\scriptsize$\pm$0.10
        &-126118.07& 1.64\scriptsize$\pm$0.81
        &170.58 & -0.17\scriptsize$\pm$0.68 \\
   25   & 884.89&-1.38\scriptsize$\pm$0.42
        & -1369.57&0.20\scriptsize$\pm$0.10
        &-126208.51& 1.57\scriptsize$\pm$0.76
        & 170.45 & -0.24\scriptsize$\pm$0.77\\  
  \midrule
  \addlinespace 
  \multicolumn{9}{c}{\large \textbf{$\tau$-PT+SA}} \\
    \addlinespace 
     \midrule
   & \multicolumn{2}{c}{MVC-2000} & \multicolumn{2}{c}{MIS-3000} & \multicolumn{2}{c}{CA-4000} & \multicolumn{2}{c}{SC-4000} \\
  \cmidrule(lr){2-3} \cmidrule(lr){4-5} \cmidrule(lr){6-7} \cmidrule(lr){8-9}
   $B$ & Obj. $\downarrow$& Gap $\downarrow$& Obj. $\downarrow$& Gap $\downarrow$& Obj. $\downarrow$& Gap $\downarrow$& Obj. $\downarrow$& Gap $\downarrow$\\
   \midrule
   5    & 883.85 & -1.50\scriptsize$\pm$0.41 
        & -1368.42 & 0.29\scriptsize$\pm$0.11
        & Infeasible (0.03\%) & - 
        & 172.80 & 1.13\scriptsize$\pm$1.08 \\
  15 (default.)  & {883.64} &{-1.52\scriptsize$\pm$0.42} 
        &-1368.98 & 0.25\scriptsize$\pm$0.11 
        &{-126212.32} &{1.57\scriptsize$\pm$0.81} 
        & 172.00 & 0.68\scriptsize$\pm$1.06 \\
   25 & 883.55 & -1.53\scriptsize$\pm$0.40
        & -1369.13 &  0.23\scriptsize$\pm$0.10
        & -126196.96 & 1.58\scriptsize$\pm$0.80
        & 171.96 & 0.65\scriptsize$\pm$1.06 \\  
  \midrule
  \addlinespace 
  \multicolumn{9}{c}{\large \textbf{$\lambda$-PT+SA}} \\
    \addlinespace 
     \midrule
   & \multicolumn{2}{c}{MVC-2000} & \multicolumn{2}{c}{MIS-3000} & \multicolumn{2}{c}{CA-4000} & \multicolumn{2}{c}{SC-4000} \\
  \cmidrule(lr){2-3} \cmidrule(lr){4-5} \cmidrule(lr){6-7} \cmidrule(lr){8-9}
  $B$ & Obj. $\downarrow$& Gap $\downarrow$& Obj. $\downarrow$& Gap $\downarrow$& Obj. $\downarrow$& Gap $\downarrow$ & Obj. $\downarrow$& Gap $\downarrow$\\
   \midrule
   5  & 885.07 & -1.36\scriptsize$\pm$0.45 
   & -1369.90 & 0.18\scriptsize$\pm$0.09
   & -125178.86 & 2.38\scriptsize$\pm$0.86 
   & 170.87 & 0.01\scriptsize$\pm$0.78 \\
   15 (default.)  & 885.78& -1.28\scriptsize$\pm$0.43 
        &{-1370.43}& {0.14\scriptsize$\pm$0.08} 
        &-125929.93 &1.79\scriptsize$\pm$0.76
        & {170.52}\  & {-0.19\scriptsize$\pm$0.83}  \\
   25 & 885.70 & -1.29\scriptsize$\pm$0.41 
        & -1370.48  &  0.14\scriptsize$\pm$0.08 
        & -126218.59 &1.57\scriptsize$\pm$0.69 
        & 170.56 & -0.17\scriptsize$\pm$0.78\\  
  \bottomrule
\end{tabular}
}
\end{table}

\subsection{Acceptance rate analysis of MLBP-$L$}
\label{app:acceptance_rate}

\begin{figure}[h]
    \centering
    \includegraphics[width=\linewidth]{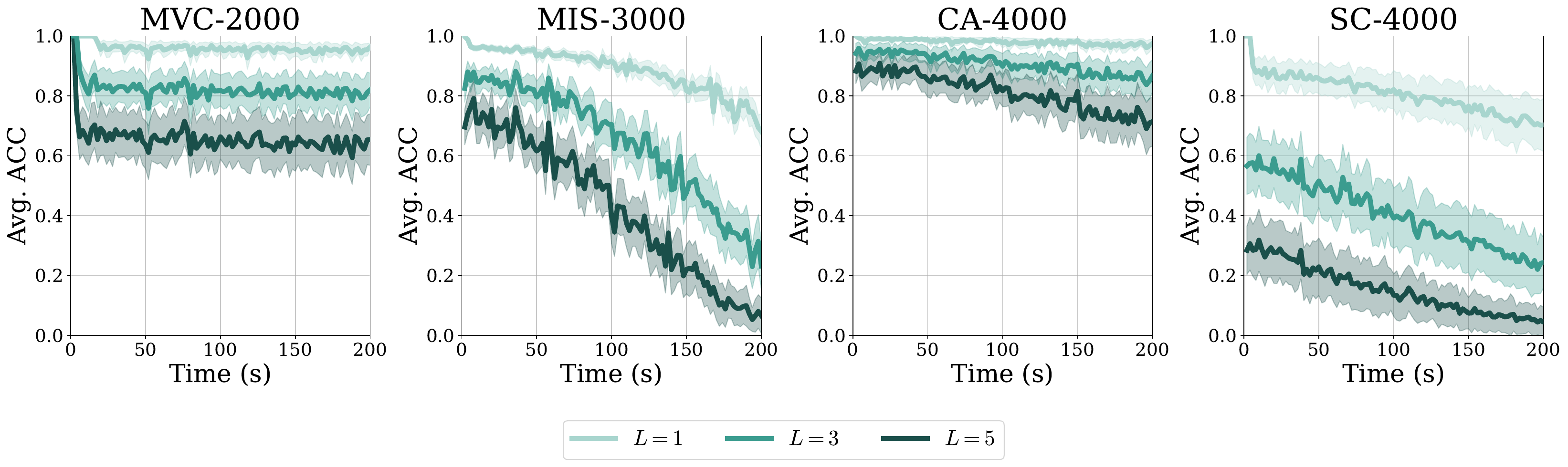}
    \caption{Average MH acceptance rate of MLBP-$L$ over time for $L=1,3,5$ across four ILP benchmarks.}
    \label{fig:mlbp_acceptance_rate}
\end{figure}

As shown in \Cref{tab:samplers}, MLBP consistently finds feasible solutions across all four tasks under both penalty formulations, whereas gradient-based methods struggle substantially. Regarding Hamming radius $L$, increasing it from MLBP-1 to MLBP-5 consistently improves the objective on MVC-2000, MIS-3000, and CA-4000, while slightly degrading performance on SC-4000. To better understand this trade-off, we further analyze MLBP under varying $L$. As illustrated in \Cref{fig:mlbp_acceptance_rate}, larger $L$ values incur greater approximation error, which in turn lowers the acceptance rate. This effect is especially severe on SC, where a single variable flip simultaneously affects many constraints and thus induces large approximation error. CA-4000 shares a similar constraint structure, but its low constraint-to-variable ratio makes this effect negligible. Overall, Hamming radius $L$ serves as a tunable knob that trades local accuracy for global coverage, while MLBP remains feasibility-robust regardless of this choice.

\section{Broader impacts}
\label{app:broader-impacts}

This work proposes a sampling-based framework for Integer Linear Programming (ILP), a general problem class with broad applications such as scheduling, routing, and resource allocation. By being training-free and solver-free, our approach lowers the barrier to entry for users without access to commercial solvers or large labeled datasets. As with any general-purpose optimization tool, the framework could in principle be applied to objectives misaligned with societal interests, but such risks are inherent to ILP itself and not specific to our method. Overall, we do not foresee any direct negative societal impacts beyond those common to optimization research.

\clearpage

\newpage

\end{document}